\documentclass{ieeeaccess}
\pdfpagewidth=\paperwidth
\pdfpageheight=\paperheight
\usepackage{cite}
\usepackage{amsmath,amssymb,amsfonts}
\usepackage{graphicx}
\usepackage[graphicx]{realboxes}
\usepackage{textcomp}

\usepackage{bm}

\usepackage{booktabs}
\usepackage{mathtools}
\usepackage{makecell}

\usepackage{multirow}
\usepackage[table,xcdraw]{xcolor}

\definecolor{higher}{RGB}{34, 139, 34} 
\definecolor{lower}{RGB}{220, 20, 60}  

\makeatletter
\AtBeginDocument{\DeclareMathVersion{bold}
\SetSymbolFont{operators}{bold}{T1}{times}{b}{n}
\SetSymbolFont{NewLetters}{bold}{T1}{times}{b}{it}
\SetMathAlphabet{\mathrm}{bold}{T1}{times}{b}{n}
\SetMathAlphabet{\mathit}{bold}{T1}{times}{b}{it}
\SetMathAlphabet{\mathbf}{bold}{T1}{times}{b}{n}
\SetMathAlphabet{\mathtt}{bold}{OT1}{pcr}{b}{n}
\SetSymbolFont{symbols}{bold}{OMS}{cmsy}{b}{n}
\renewcommand\boldmath{\@nomath\boldmath\mathversion{bold}}}
\makeatother

\def\BibTeX{{\rm B\kern-.05em{\sc i\kern-.025em b}\kern-.08em
    T\kern-.1667em\lower.7ex\hbox{E}\kern-.125emX}}

\begin{document}
\history{Received 18 November 2025, accepted 9 January 2026, date of publication 19 January 2026, date of current version 29 January 2026.}
\doi{Digital Object Identifier 10.1109/ACCESS.2026.3655392\\Published in IEEE Access, vol. 14, pp. 14709--14721, 2026.}
\vol{14}\year{2026}
\pubid{\parbox{0.73\textwidth}{\centering\fontsize{5.5}{6.5}\selectfont\textcopyright\ 2026 The Authors. This work is licensed under a Creative Commons Attribution 4.0 License.\\For more information, see https://creativecommons.org/licenses/by/4.0/}}

\title{InterHier: Learning Interconnected Hierarchical Semantics for Open-Vocabulary Object Detection}
\author{\uppercase{Yeong-Jin Kim}, \uppercase{Ho-Joong Kim}, and \uppercase{Seong-Whan Lee}, 
\IEEEmembership{Fellow, IEEE}}

\address[1]{Department of Artificial Intelligence, Korea University, Seoul 02841, Republic of Korea}

\tfootnote{This work was supported in part by the Institute of Information and Communications Technology Planning and Evaluation (IITP) funded by Korean Government (MSIT) through the Artificial Intelligence Graduate School Program (Korea University) under Grant RS-2019-II190079, and in part by the Information Technology Research Center (ITRC) under Grant IITP-2025-RS-2024-00436857.}

\markboth
{Y.-J. Kim \headeretal: InterHier: Learning Interconnected Hierarchical Semantics for Open-Vocabulary Object Detection}
{Y.-J. Kim \headeretal: InterHier: Learning Interconnected Hierarchical Semantics for Open-Vocabulary Object Detection}

\corresp{Corresponding author: Seong-Whan Lee (e-mail: sw.lee@korea.ac.kr)}

\begin{abstract}
Open-vocabulary object detection (OVD) aims to localize and classify objects from arbitrary categories. These categories are specified through textual input and are not limited to a predefined set. The existing methods explicitly utilize hierarchical semantic representations of super-/sub-categories to establish semantic relationships between base categories and unseen novel categories. These methods rely on a fixed connector such as ``, which is a'', and this fixed connector is placed between each adjacent super-/sub-category to integrate them together. However, these methods are not an optimal solution because these relationships rely on hand-crafted connectors. To address this issue, we propose interconnected hierarchical semantic representations (InterHier). InterHier utilizes a prepended learnable context to globally guide the interpretation of prompts containing hierarchical relationships. InterHier operates in two main stages. In the first stage, InterHier constructs the hierarchy-aware prompt by integrating super-/sub-categories and then prepending a learnable context. In the second stage, InterHier optimizes this learnable context to align the visual region embeddings and textual embeddings. InterHier demonstrates consistently improved performance over methods that rely on fixed connectors. Moreover, InterHier offers high versatility and allows for seamless integration into existing OVD models. On OVD benchmarks, InterHier achieves competitive performance against other state-of-the-art methods.
\end{abstract}

\begin{keywords}
Hierarchy-aware prompt, learnable context, object detection, open-vocabulary, prompt learning, semantic alignment, semantic hierarchy
\end{keywords}

\titlepgskip=-21pt

\maketitle

\section{Introduction}
\label{sec:introduction}
Recent advancements in computer vision~\cite{he2016deep, dosovitskiy2021image, kim2024te, wang2025weakly, 7024902, peng2025d, 10004558, 711986, wang2025large, ROH2007931, peng2025navigscene, 9658533, su2025real, 10371296, WANG2025130571, lee2024text, WANG2026110144} have improved a wide range of visual understanding tasks, such as image classification, object detection, and semantic segmentation. In particular, the development of vision-language models (VLMs)~\cite{radford2021learning, jia2021scaling, li2022blip} has enabled effective joint learning of visual and textual information. This joint learning has led to improvements in model expressiveness and generalization across multi-modal reasoning tasks. These tasks include image captioning~\cite{vinyals2015show, li2020oscar, cheng2025caparena}, visual question answering~\cite{antol2015vqa, goyal2017making, anderson2018bottom}, and vision-text retrieval~\cite{lee2018stacked, chen2020uniter, kim2021vilt}. Building upon these developments, open-vocabulary object detection (OVD) has emerged as a crucial task. OVD aims to detect objects from arbitrary categories that are specified through textual input and are not limited to a predefined set. This capability is enabled by VLMs such as CLIP~\cite{radford2021learning} and ALIGN~\cite{jia2021scaling}. These models are trained on extensive image-text datasets to establish a semantic alignment between visual and textual embeddings.

While the semantic alignment provides capabilities for image-level recognition, this alignment often struggles to capture the complex relationships between object categories. To further improve the recognition of diverse and unseen categories, the existing methods~\cite{huang2024openvocabulary, liu2024shine} explore hierarchical semantic structures. For instance, DetLH~\cite{huang2024openvocabulary} leverages a language hierarchy to structure semantic relationships between categories. DetLH expands weak image-level labels to generate more accurate pseudo-labels for the detector. Additionally, SHiNe~\cite{liu2024shine} constructs hierarchy-aware prompts by using a fixed connector such as ``, which is a'', to sequentially integrate categories from the coarsest level to the finest level. The coarsest level represents a general super-category such as ``vehicle'', while the finest level represents a specific sub-category such as ``scooter''. Through this integration, SHiNe is designed to improve the semantic relationships between super-/sub-categories. However, despite this sequential integration of categories, these methods are not an optimal solution because these prompts rely on hand-crafted connectors.

\begin{figure}[t!]
    \centering
    \includegraphics[width=\linewidth]{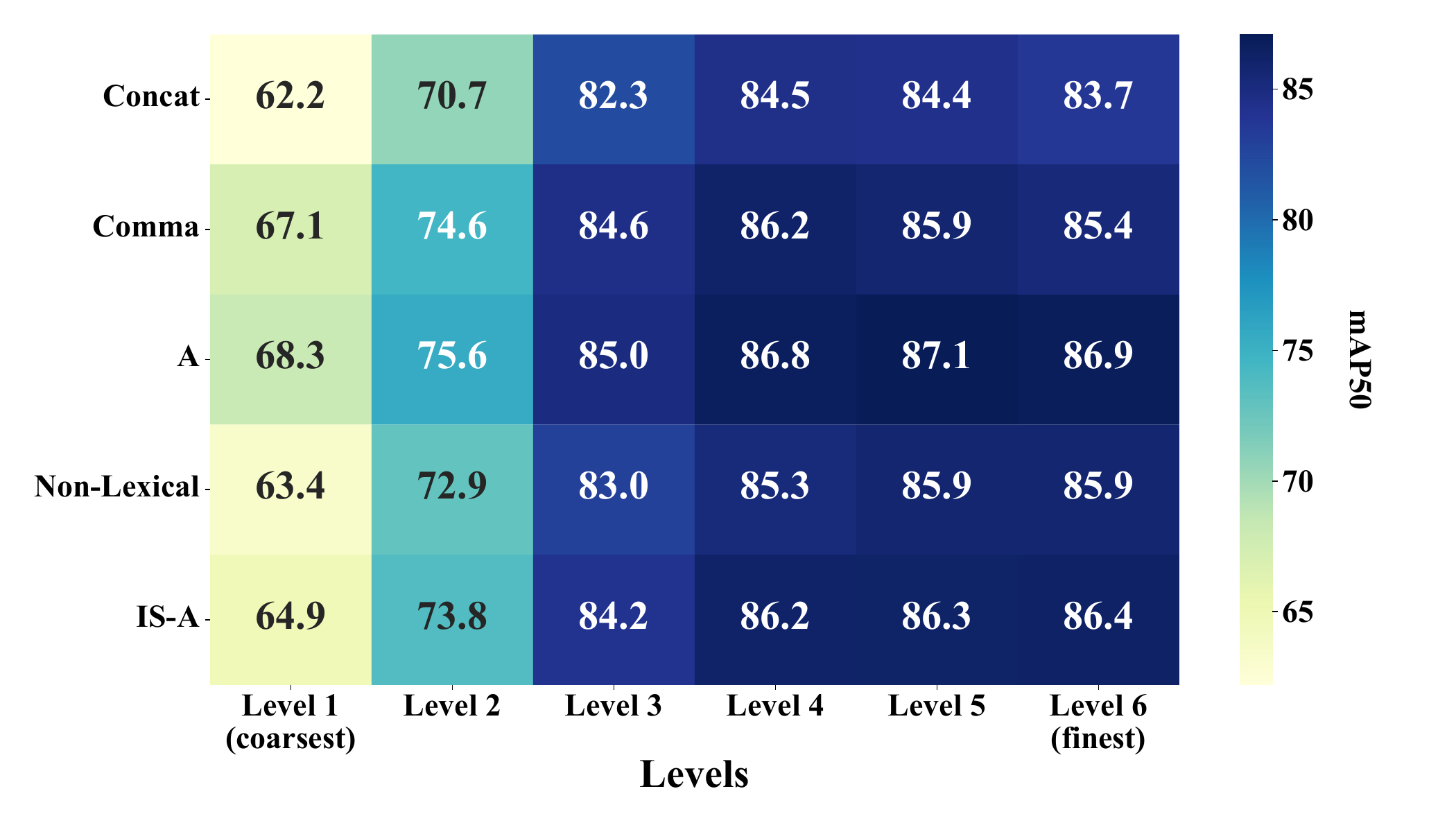}
    \includegraphics[width=0.9\linewidth]{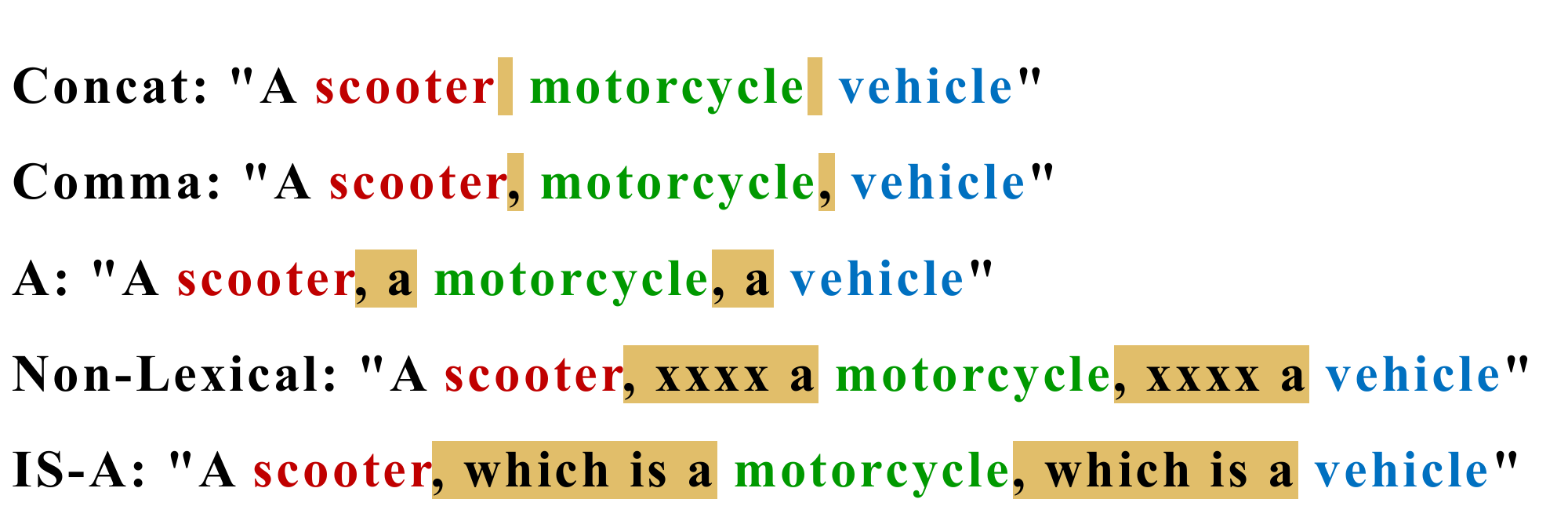}
\caption{Comparative analysis of different fixed connectors for constructing hierarchy-aware prompts on iNatLoc~\cite{cole2022label}: (top) Performance comparison across six levels of semantic granularity, using mean average precision (mAP) computed at an intersection-over-union (IoU) threshold of 0.5 (mAP50) as the evaluation metric. The x-axis illustrates the granularity levels, from the coarsest (Level 1) to the finest (Level 6). The y-axis represents the five different fixed connector evaluated; (bottom) Detailed examples of how each fixed connector is used to integrate super-/sub-category relationships. The examples shown are based on the level 3 hierarchy (scooter, motorcycle, vehicle). The connectors are defined as: Concat: directly appends a string of one class to the end of a string of another class. Comma: separates two category names with a comma. A: separates two category names with ``, a''. Non-Lexical: separates two category names with non-lexical. IS-A: separates two category names with ``, which is a''.}
\label{fig:1}
\end{figure}

To demonstrate this issue, Fig. \ref{fig:1} presents a comparative analysis evaluating different fixed connectors used to integrate these hierarchical relationships across the six label granularity levels on iNatLoc~\cite{cole2022label}. First, the Comma connector clearly separates each category, showing improved performance over the Concat connector, which lacks this separation. This result is attributed to the Concat connector creating a single, grammatically unnatural phrase. Additionally, the simpler A connector demonstrates a more concise representation of the categories than the IS-A connector~\cite{liu2024shine}. The verbose IS-A connector includes non-visual glue words such as ``which'' and ``is''. These extra words require the VLM to process unnecessary grammatical structures that have no direct correspondence with the visual features. Furthermore, this formal grammatical structure creates a mismatch with the pre-training data of the model, which largely consists of concise captions rather than verbose structured sentences. Finally, the Non-Lexical connector that does not encode hierarchical semantic meaning exhibits insignificant performance differences from the IS-A connector. These results suggest that the inherent limitations of using hand-crafted connectors prevent these methods from being an optimal solution.

In this paper, we propose interconnected hierarchical semantic representations (InterHier), a novel method that enhances the representation of hierarchical semantic relationships. Instead of relying on fixed connectors, InterHier utilizes a prepended learnable context that globally guides the interpretation of the entire prompt. This global learnable context enables the self-attention mechanism of the text encoder to capture the complex hierarchical relationships between all categories. The framework of InterHier proceeds in two main stages. In the first stage, InterHier constructs the hierarchy-aware prompt by integrating super-/sub-categories and then prepending a learnable context. In the second stage, InterHier optimizes this learnable context to align the visual region embeddings and their corresponding textual embeddings. InterHier demonstrates consistently improved performance over methods that rely on fixed connectors. Moreover, InterHier is designed to seamlessly integrate into existing OVD models and allows its adoption across diverse hierarchical structures. To this end, we validate the effectiveness of InterHier through extensive experiments on multiple OVD benchmarks. The results on iNatLoc and the existing few-shot object detection (FSOD) benchmark~\cite{fan2020few} show that InterHier consistently improves hierarchical semantic representations. These findings highlight the adaptability of InterHier across varying levels of semantic granularity and supervision settings. Furthermore, InterHier achieves competitive performance on the open-vocabulary LVIS (OV-LVIS)~\cite{gupta2019lvis} and the open-vocabulary COCO (OV-COCO) benchmarks~\cite{lin2014microsoft}. These experiments demonstrate its effectiveness against other state-of-the-art methods. Additionally, we conduct a qualitative evaluation to verify the effectiveness of InterHier. These results confirm that InterHier enhances the discriminative representations that fixed connectors fail to provide.

In summary, our main contributions are as follows:
\begin{itemize}
\item We propose InterHier, which utilizes a prepended learnable context to globally guide the interpretation of hierarchical semantic relationships
\item InterHier is designed to seamlessly integrate into existing OVD models, demonstrating its high versatility and robustness across diverse hierarchical structures.
\item InterHier achieves superior performance over various state-of-the-art methods on OVD benchmarks: iNatLoc, FSOD, OV-LVIS, OV-COCO.
\end{itemize}

\section{RELATED WORKS}
\label{sec:related_works}

\subsection{Open-Vocabulary Object Detection}
OVD~\cite{WuPAMI23TowardsOpenVocabularyLearning, zhu2024survey, LiWWLLY25} focuses on localizing and classifying object instances from a set of categories unseen during training. This task distinguishes itself from standard zero-shot detectors~\cite{tan2021survey} by associating visual region embeddings at the region level with textual embeddings. These embeddings are obtained from VLMs~\cite{radford2021learning, jia2021scaling, li2022blip} that have been trained to associate images with corresponding textual descriptions. This association is performed by calculating the cosine similarity between the visual embedding of a region and the textual embedding of a potential class name. Region-text alignment methods~\cite{zhou2022detecting, zhong2022regionclip, ma2023codet, kim2024region, xiao2025textregion} have been developed to mitigate the disparity between visual detection tasks and textual class descriptions. These methods enhance zero-shot object detection capabilities and generalization to unseen novel classes. Building upon these region-text alignment methods, InterHier utilizes a pre-trained OVD model as its base detector and enhances its classification performance.

\subsection{Semantic Hierarchy}
Semantic hierarchy~\cite{fellbaum1998wordnet,van2018inaturalist} organizes conceptual labels into a relational structure. This structure often takes the form of a tree-like taxonomy~\cite{wu2005learning}. Such hierarchies have been leveraged in computer vision to enhance various tasks~\cite{deng2010does,frome2013devise,goodman2001classes,bertinetto2020making}. Early works used the structure of WordNet~\cite{fellbaum1998wordnet} to define a semantic distance between categories. This method aided knowledge transfer in zero-shot learning and improved object classification. Recent work on large-scale datasets with built-in hierarchies~\cite{cole2022label} shows that appropriate label granularity improves performance when training models for weakly supervised object localization. Building on this, SHiNe~\cite{liu2024shine} applies a semantic hierarchy nexus classifier to pre-trained region-text aligned OVD detectors. This method relies on hand-crafted connectors that are not optimal for representing the relationships between super-/sub-categories. In contrast, InterHier utilizes a prepended learnable context to improve the representation of hierarchical semantic relationships.

\subsection{Prompt Learning}
Prompt learning~\cite{liu2023pre} has been studied as a learning method designed to leverage the capabilities of pretrained large language models~\cite{devlin2019bert, radford2019language, raffel2020exploring} or VLMs~\cite{radford2021learning, jia2021scaling, li2022blip}. While conventional methods involved manually designed fixed prompts, recent research has explored methods that incorporate prompts themselves as learnable parameters within the model. The prompt tuning methods~\cite{zhou2022learning, ju2022prompting, zhou2022conditional} incorporate learnable token embeddings into textual prompts. These methods enable task-specific fine-tuning for downstream tasks. CoOp~\cite{zhou2022learning} introduces context optimization, which formulates prompts as continuous embeddings. These embeddings are updated through end-to-end training to automate the prompt engineering process. This automated process facilitates the adaptation of pretrained VLMs for downstream tasks. InterHier applies prompt learning to hierarchical semantic representation by prepending a learnable context. This learnable context is modeled as a continuous representation that is optimized to globally guide the interpretation of the prompt.

\begin{figure*}[!t]
    \centering
    \includegraphics[width=\linewidth]{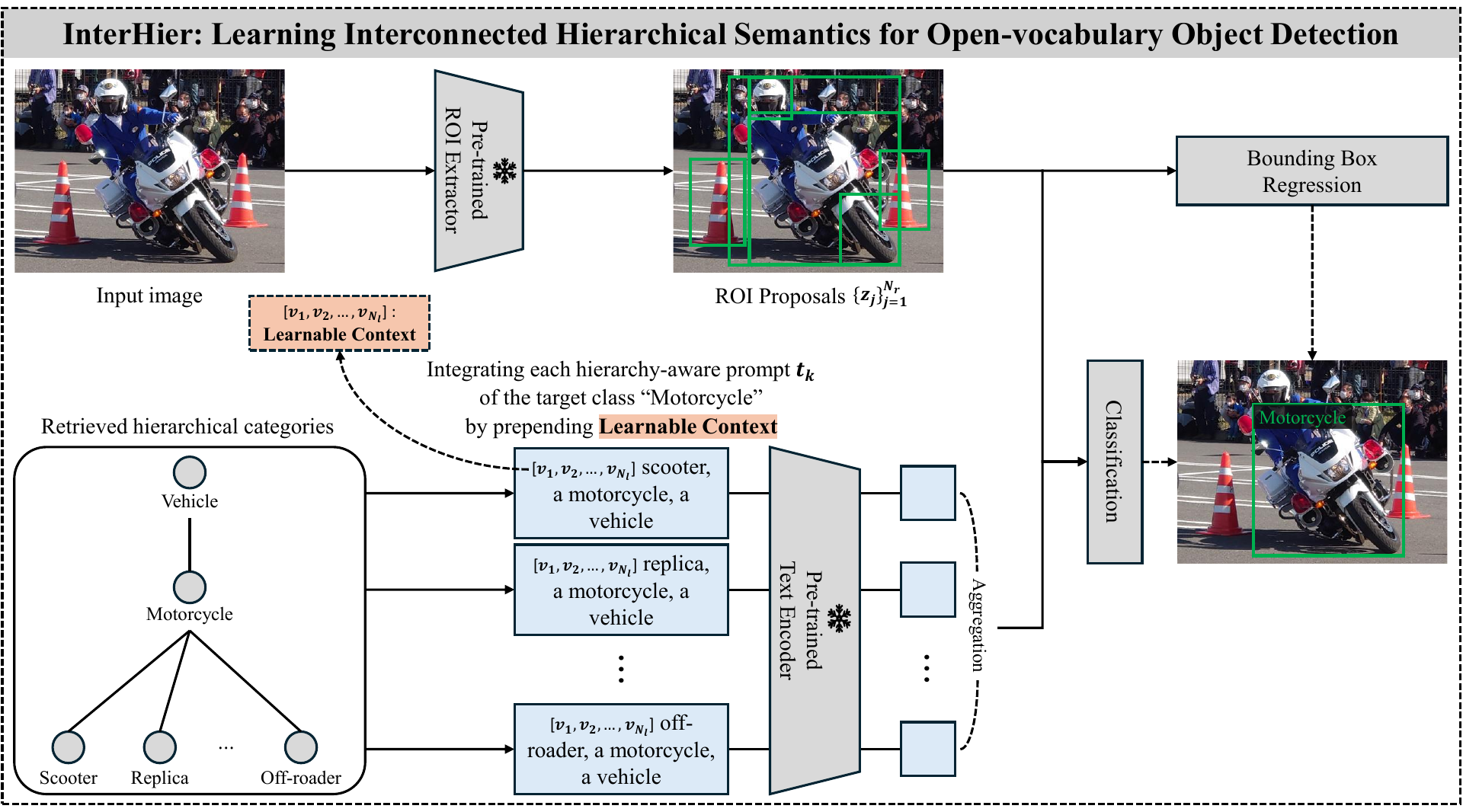}
    \caption{
    Overview of the InterHier. The main idea is to construct hierarchy-aware prompt embeddings by prepending a set of learnable context vectors to prompts that represent the semantic hierarchy. These vectors integrate retrieved hierarchical categories into structured prompts and are optimized by minimizing the semantic alignment loss between visual and textual embeddings. The constructed prompts are encoded via a frozen text encoder, and the resulting embeddings are aggregated to form the hierarchy-aware classifier. The classifier is then used to assign semantic labels to the visual embeddings of the ROI proposals, which are combined with refined bounding box coordinates to yield the final detection.}
\label{fig:2}
\end{figure*}

\section{METHOD}
\label{sec:method}
In this section, we describe InterHier, a method designed to enhance hierarchical semantic relationships in the OVD task. The framework of InterHier operates in two main stages. In the first stage, InterHier constructs the hierarchy-aware prompt by integrating super-/sub-categories and then prepending a learnable context. In the second stage, InterHier optimizes this learnable context to align the visual region embeddings and their corresponding textual embeddings. An overview is illustrated in Fig. \ref{fig:2}.

\subsection{Preliminary}
\subsubsection{Problem formulation}
The primary goal of OVD is to detect objects from arbitrary categories within an image using textual input. These categories are not limited to a predefined set. This task is performed in a zero-shot manner, without requiring any additional model training. Given an input image, a model aims to identify all foreground objects by estimating bounding box coordinates and corresponding class labels, represented as:

\begin{equation}
\{{b}_m, d_m\}_{m=1}^{N_d},
\end{equation}
where $N_d$ is the number of detected objects. Each ${b}_m \in \mathbb{R}^4$ denotes the bounding box coordinates for the $m$-th object. The corresponding class label $d_m$ is selected from $\mathcal{C}^{\mathrm{test}}$, which is the user-specified vocabulary set used at test time.

\subsubsection{Open-vocabulary object detector}
Predominant OVD detectors~\cite{zhou2022detecting, ma2023codet} are built upon a two-stage detection framework. This process begins with an input image. In the first stage, a region proposal network (RPN) processes the image to construct a set of $N_r$ candidate region proposals. This step yields a collection of region of interest (RoI) feature embeddings, which we denote as $\{z_j\}_{j=1}^{N_r}$. In the second stage, each RoI feature embedding $z_j$ is processed by two separate heads for box regression and classification. The regression head refines the bounding box coordinates, while the classification head assigns a class label. For open-vocabulary capability, the weights of the classifier are derived from textual embeddings of the user-defined classes $\mathcal{C}^{\mathrm{test}}$, which are obtained from a pre-trained VLM. The classification is determined by computing the cosine similarity between each visual RoI embedding $z_j$ and the textual embedding $z_c$ for every candidate class. The target class yielding the highest similarity score is assigned to the object.

\subsubsection{Querying the semantic hierarchy}
To construct the hierarchy-aware classifier weights, our process begins by utilizing the semantic hierarchy, which provides structured relationships between object categories. In this work, we derive this hierarchy directly from the ground-truth class taxonomies provided by the datasets~\cite{cole2022label, fan2020few}. This hierarchy is utilized to process each target class $c$ within the test vocabulary $\mathcal{C}^{\mathrm{test}}$. The composition of this vocabulary $\mathcal{C}^{\mathrm{test}}$ is dependent on the selected semantic granularity level.

For each target class $c$ within the test vocabulary $\mathcal{C}^{test}$, our process begins by identifying all finest-level categories that are sub-categories of $c$ within the given taxonomy. We construct a hierarchical series for each of these identified finest-level categories. A valid traversal is defined as the process of constructing an ordered list of categories. This construction starts from a specific finest-level category. The traversal proceeds upwards by following its super-category connections. The list is complete when the traversal reaches the target class $c$, which is also included in the series. For instance, let the target class $c$ be ``vehicle'' at level 3. One of its finest-level sub-categories is ``scooter''. The valid traversal starting from ``scooter'' constructs the hierarchical series (\texttt{scooter}, \texttt{motorcycle}, \texttt{vehicle}). This process is repeated for all other finest-level sub-categories of ``vehicle''. This procedure yields a set of hierarchical series.

Each hierarchical series is subsequently converted into a hierarchy-aware prompt, defined as $t_k$. The construction of the hierarchy-aware prompt involves joining these category names with a specific textual connector. As shown in Fig. \ref{fig:1}, the verbose IS-A connector~\cite{liu2024shine} includes non-visual glue words such as ``which'' and ``is''. These words require the VLM~\cite{radford2021learning, jia2021scaling} to process unnecessary grammatical structures that have no direct correspondence with the visual features. Therefore, we adopt the simpler A connector for integrating the adjacent categories. For instance, the series of category names (\texttt{scooter}, \texttt{motorcycle}, \texttt{vehicle}) is converted into the prompt \texttt{scooter, a motorcycle, a vehicle}. The resulting prompts provide the structured input for InterHier.

\subsection{InterHier}

\subsubsection{Hierarchical Prompt Learning}
To improve the representation of these hierarchical relationships, InterHier introduces a set of learnable context $v$, which is composed of a set of individual context vectors. The purpose of this learnable context is to globally guide the interpretation of the hierarchy-aware prompts. The learnable context $v$ is defined as follows:

\begin{equation}
v = \text{Concat}(v_1, v_2, \dots, v_{N_l}),
\label{eq:2}
\end{equation}
where $N_l$ denotes the number of context vectors.

To apply this learnable context $v$, we construct an input vector $x_k$. We tokenize the hierarchy-aware prompt $t_k$ using the $\text{Tokenize}(\cdot)$ function adopted from CLIP~\cite{radford2021learning}, which converts the text into a sequence of token embeddings. The learnable context vectors $v$ are prepended to the beginning of this sequence. The input vector $x_k$ is specified as follows:

\begin{equation}
x_k = \text{Concat}(v, \text{Tokenize}(t_k)),
\end{equation}
where $x_k$ is the concatenated sequence of the learnable context and the token embeddings. This input vector $x_k$ serves as a rich, hierarchy-aware representation for a single hierarchical series. After constructing an input vector $x_k$ for each finest-level category, we group the resulting vectors into a set. The set of all input vectors $S_c$ is represented as follows:

\begin{equation}
    S_c = \left\{ x_k \mid k \in  
        \begin{array}{c}
            \text{ finest-level category index} \\
            \text{of the target class } c 
        \end{array} 
    \right\}.
\end{equation}

Each input vector $x_k$ is passed through the pre-trained and frozen text transformer from CLIP, which is denoted as $\text{TextEncoder}(\cdot)$. To construct a single unified representation, we aggregate these resulting textual embeddings. Specifically, we compute the arithmetic mean of all the constructed textual embeddings. The final hierarchy-aware textual embeddings $z_c$, which are defined as follows:

\begin{equation}
    z_c = \frac{1}{n(S_c)}\sum_{x_k \in S_c}{\text{TextEncoder}(x_k)}. 
\end{equation}

The final embedding $z_c$ serves as the hierarchy-aware classifier weight for the target class $c$. The effectiveness of our method stems from the interaction between the learnable context and the self-attention mechanism of the text encoder. The prepended learnable context guides the model to interpret the sequence as a hierarchy. The self-attention layers consider all tokens in the sequence simultaneously. Therefore, the self-attention layers leverage this initial global context to infer the relationships between all category names.

\subsubsection{Training and Inference}
In the training phase, we optimize the learnable context vectors $v$. This optimization is achieved by aligning the resulting class-level textual embedding $z_c$ with the corresponding visual RoI embeddings $z_j$. Specifically, InterHier employs a contrastive loss function, similar to the one used in CLIP. For a given visual embedding $z_j$ belonging to a ground-truth class $c^+$, we design the loss to maximize its cosine similarity with the matching textual embedding $z_{c^+}$. Simultaneously, the loss minimizes the similarity between $z_j$ and the embeddings of all other unrelated classes at that level. The total loss for the $N_r$ region proposals is the average of the individual losses, formulated as follows:

\begin{equation}
    \mathcal{L} = -\frac{1}{N_{r}} \sum_{j=1}^{N_{r}} \log\left(\frac{\exp(\text{sim}(z_j, z_{c^+}) / \tau)}{\sum_{i=1}^{N_c} \exp(\text{sim}(z_j, z_{c_i}) / \tau)}\right),
\end{equation}
where $\text{sim}(\cdot,\cdot)$ denotes cosine similarity, $N_c$ is the total number of training classes, $z_{c_i}$ is the textual embedding for the $i$-th class, and $\tau$ is a temperature hyperparameter. During this process, the visual encoder and detection head of the base model remain frozen. This prompt-tuning ensures that only the context vectors are updated. This method allows for efficient optimization while leveraging the pre-trained representations of the base detector.

At inference time, the optimized context vectors are prepended to the hierarchy-aware prompts. These prompts are encoded via the frozen $\text{TextEncoder}(\cdot)$ to construct the textual embeddings $z_c$ for all classes at the desired evaluation levels. This set of embeddings forms the new classifier weight matrix. The classification of a region proposal with visual embedding $z_j$ is then performed by identifying the class with the highest cosine similarity, as follows:

\begin{equation}
    \hat{c}_j = \underset{c \in \mathcal{C}^{\mathrm{test}}}{\arg\max} \left( \text{sim}(z_j, z_c) \right),
\end{equation}
where $\hat{c}_j$ is the predicted class label for the $j$-th region proposal. The design of our method allows seamless integration into existing OVD models without requiring any architectural modifications to the base detector.

\section{EXPERIMENTS}
\label{sec:experiments}

\subsection{Datasets}

\begin{table}[h!]
    \centering
    \footnotesize
    \renewcommand{\arraystretch}{1.1}
    \caption{
    Evaluation dataset descriptions of iNatLoc~\cite{cole2022label} and FSOD~\cite{fan2020few}. Label granularity ranges from the coarsest (C) level to the finest (F) level.}
    \begin{tabular}{c|ccl|ccl}
         \toprule
         \multirow{2.5}{*}{\rotatebox[origin=c]{90}{Gran}}
         & \multicolumn{3}{c|}{iNatLoc} 
         & \multicolumn{3}{c}{FSOD}\\

         \cmidrule(lr){2-4}
         \cmidrule(lr){5-7}

         & Level &  Classes & \multicolumn{1}{c|}{Sample} 
         & Level &  Classes & \multicolumn{1}{c}{Sample}\\

         \midrule

         \multirow{6}{*}{\rotatebox[origin=c]{90}{C $\xrightarrow{\hspace{1.2cm}}$ F}}
         & 6 & 500 & Felis catus
         & \multirow{2}{*}{3} & \multirow{2}{*}{200} & \multicolumn{1}{c}{\multirow{2}{*}{Beer}} \\
         
         & 5 & 317 & Felis
         &  &  & \\
         
         & 4 & 184 & Felidae
         & \multirow{2}{*}{2} & \multirow{2}{*}{46} & \multicolumn{1}{c}{\multirow{2}{*}{Drink}} \\
         
         & 3 & 64  & Carnivora
         &  &  & \\
         
         & 2 & 18  & Mammalia
         & \multirow{2}{*}{1} & \multirow{2}{*}{15} & \multicolumn{1}{c}{\multirow{2}{*}{Liquid}} \\
         
         & 1 & 5   & Chordata
         &  &  & \\
         \bottomrule
    \end{tabular}
    \label{tab:1}
\end{table}

\subsubsection{iNatLoc and FSOD}
iNatLoc~\cite{cole2022label} and FSOD~\cite{fan2020few} offer ground-truth hierarchical annotations to support multi-level label granularity evaluation. iNatLoc is used for weakly supervised object localization and features a six-level biological taxonomy. This structure is organized from the coarsest level (``Phylum'') to the finest level (``Species''). As the hierarchy descends from coarser to finer levels, the intra-class variation decreases significantly. For instance, categories at the coarsest level contain broad, diverse objects, whereas categories at the finest level represent specific and visually consistent objects. For evaluation, iNatLoc provides bounding box annotations for its validation split. FSOD is a dataset curated for few-shot object detection, constructed by combining Open Images~\cite{kuznetsova2020open} and ImageNet~\cite{deng2009imagenet}. FSOD features a two-level hierarchical structure. For extended analysis, we use FSOD’s test split and directly adopt the three-level hierarchical structure manually constructed in SHiNe~\cite{liu2024shine}. Tab. \ref{tab:1} presents the number of label hierarchy levels and their corresponding category distributions for both datasets, supplemented with samples to highlight differences in semantic granularity.

\begin{table*}[!ht]
    \centering
    \small
    \renewcommand{\arraystretch}{1.5}
    \caption{
    Detection results are evaluated on iNatLoc~\cite{cole2022label} across six label granularity levels, ranging from the coarsest (C) level to the finest (F) level, using ground-truth hierarchies. InterHier is directly integrated into the baseline without any architectural changes. ResNet-50~\cite{he2016deep} and Swin-B~\cite{liu2021swin} backbones are compared. Each backbone is evaluated under various supervision settings: LVIS~\cite{gupta2019lvis}, LVIS + IN-L~\cite{deng2009imagenet}, LVIS + IN-21k~\cite{deng2009imagenet}, and LVIS \& COCO~\cite{lin2014microsoft} + IN-21k. mAP50 is reported.}
    \resizebox{\linewidth}{!}{
    \begin{tabular}{ll|ll|ll|ll|ll|ll|ll|}
        \toprule
        &\multicolumn{1}{c}{}
        & \multicolumn{4}{c}{ResNet-50 Backbone}
        & \multicolumn{8}{c}{Swin-B Backbone} \\

        \cmidrule(lr){1-2}
        \cmidrule(r){3-6}
        \cmidrule(r){7-14}

        \multicolumn{2}{c}{iNatLoc}
        & \multicolumn{2}{|c}{LVIS} 
        & \multicolumn{2}{c|}{LVIS + IN-L}
        & \multicolumn{2}{c}{LVIS} 
        & \multicolumn{2}{c}{LVIS + IN-L} 
        & \multicolumn{2}{c}{LVIS + IN-21k} 
        & \multicolumn{2}{c}{LVIS \& COCO + IN-21k} \\

        \cmidrule(lr){3-4}
        \cmidrule(lr){5-6}
        \cmidrule(lr){7-8}
        \cmidrule(lr){9-10}
        \cmidrule(lr){11-12}
        \cmidrule(lr){13-14}

        \multirow{1}{*}{\rotatebox[origin=c]{90}{Gran}} & \multirow{1}{*}{\rotatebox[origin=c]{90}{Lev}}
         
        & \multicolumn{1}{c}{SHiNe} & \multicolumn{1}{c|}{InterHier}
        & \multicolumn{1}{c}{SHiNe} & \multicolumn{1}{c|}{InterHier}
        & \multicolumn{1}{c}{SHiNe} & \multicolumn{1}{c|}{InterHier}
        & \multicolumn{1}{c}{SHiNe} & \multicolumn{1}{c|}{InterHier}
        & \multicolumn{1}{c}{SHiNe} & \multicolumn{1}{c|}{InterHier}
        & \multicolumn{1}{c}{SHiNe} & \multicolumn{1}{c}{InterHier} \\

        \cmidrule(lr){3-4}
        \cmidrule(lr){5-6}
        \cmidrule(lr){7-8}
        \cmidrule(lr){9-10}
        \cmidrule(lr){11-12}
        \cmidrule(lr){13-14}
        
        \multirow{6}{*}{\rotatebox[origin=c]{90}{C $\xrightarrow{\hspace{2.1cm}}$ F}}

        & L6
        & \multicolumn{1}{c}{48.4} & \multicolumn{1}{c|}{\bf54.3({\color{higher}+5.9})}
        & \multicolumn{1}{c}{57.1} & \multicolumn{1}{c|}{\bf60.8({\color{higher}+3.7})}
        & \multicolumn{1}{c}{76.7} & \multicolumn{1}{c|}{\bf80.9({\color{higher}+4.2})}
        & \multicolumn{1}{c}{83.8} & \multicolumn{1}{c|}{\bf87.3({\color{higher}+3.5})}
        & \multicolumn{1}{c}{86.3} & \multicolumn{1}{c|}{\bf88.6({\color{higher}+2.3})}
        & \multicolumn{1}{c}{86.4} & \multicolumn{1}{c}{\bf89.1({\color{higher}+2.7})} \\
								
        & L5 
        & \multicolumn{1}{c}{49.4} & \multicolumn{1}{c|}{\bf56.6({\color{higher}+7.2})}
        & \multicolumn{1}{c}{59.0} & \multicolumn{1}{c|}{\bf63.0({\color{higher}+4.0})}
        & \multicolumn{1}{c}{77.7} & \multicolumn{1}{c|}{\bf82.9({\color{higher}+5.2})}
        & \multicolumn{1}{c}{84.8} & \multicolumn{1}{c|}{\bf87.9({\color{higher}+3.1})}
        & \multicolumn{1}{c}{86.8} & \multicolumn{1}{c|}{\bf88.8({\color{higher}+2.0})}
        & \multicolumn{1}{c}{86.3} & \multicolumn{1}{c}{\bf89.0({\color{higher}+2.7})} \\
 										
        & L4 
        & \multicolumn{1}{c}{51.5} & \multicolumn{1}{c|}{\bf58.3({\color{higher}+6.8})}
        & \multicolumn{1}{c}{61.4} & \multicolumn{1}{c|}{\bf64.6({\color{higher}+3.2})}
        & \multicolumn{1}{c}{76.6} & \multicolumn{1}{c|}{\bf83.2({\color{higher}+6.6})}
        & \multicolumn{1}{c}{84.4} & \multicolumn{1}{c|}{\bf88.4({\color{higher}+4.0})}
        & \multicolumn{1}{c}{87.7} & \multicolumn{1}{c|}{\bf89.5({\color{higher}+1.8})}
        & \multicolumn{1}{c}{86.2} & \multicolumn{1}{c}{\bf89.0({\color{higher}+2.8})} \\
								
        & L3 
        & \multicolumn{1}{c}{56.5} & \multicolumn{1}{c|}{\bf62.4({\color{higher}+5.9})}
        & \multicolumn{1}{c}{65.3} & \multicolumn{1}{c|}{\bf68.2({\color{higher}+2.9})}
        & \multicolumn{1}{c}{75.3} & \multicolumn{1}{c|}{\bf82.4({\color{higher}+7.1})}
        & \multicolumn{1}{c}{84.4} & \multicolumn{1}{c|}{\bf86.3({\color{higher}+1.9})}
        & \multicolumn{1}{c}{86.9} & \multicolumn{1}{c|}{\bf89.3({\color{higher}+2.4})}
        & \multicolumn{1}{c}{84.2} & \multicolumn{1}{c}{\bf87.2({\color{higher}+3.0})} \\
				
        & L2 
        & \multicolumn{1}{c}{45.0} & \multicolumn{1}{c|}{\bf51.5({\color{higher}+6.5})}
        & \multicolumn{1}{c}{53.7} & \multicolumn{1}{c|}{\bf56.2({\color{higher}+2.5})}
        & \multicolumn{1}{c}{62.1} & \multicolumn{1}{c|}{\bf70.2({\color{higher}+8.1})}
        & \multicolumn{1}{c}{73.1} & \multicolumn{1}{c|}{\bf76.6({\color{higher}+3.5})}
        & \multicolumn{1}{c}{78.1} & \multicolumn{1}{c|}{\bf82.6({\color{higher}+4.5})}
        & \multicolumn{1}{c}{73.8} & \multicolumn{1}{c}{\bf77.6({\color{higher}+3.8})} \\
                                            
        & L1
        & \multicolumn{1}{c}{33.6} & \multicolumn{1}{c|}{\bf39.9({\color{higher}+6.3})}
        & \multicolumn{1}{c}{43.3} & \multicolumn{1}{c|}{\bf45.9({\color{higher}+2.6})}
        & \multicolumn{1}{c}{49.7} & \multicolumn{1}{c|}{\bf65.2({\color{higher}+15.5})}
        & \multicolumn{1}{c}{59.4} & \multicolumn{1}{c|}{\bf69.9({\color{higher}+10.5})}
        & \multicolumn{1}{c}{70.3} & \multicolumn{1}{c|}{\bf78.9({\color{higher}+8.6})}
        & \multicolumn{1}{c}{64.9} & \multicolumn{1}{c}{\bf75.2({\color{higher}+10.3})} \\
         
        \bottomrule
    \end{tabular}
    }
    \label{tab:2}
\end{table*}
\begin{table*}[h!]
    \centering
    \small
    \renewcommand{\arraystretch}{1.5}
    \caption{
    Detection results are evaluated on FSOD~\cite{fan2020few} across three label granularity levels, ranging from the coarsest (C) level to the finest (F) level, using ground-truth hierarchies. InterHier is directly integrated into the baseline without any architectural changes. ResNet-50~\cite{he2016deep} and Swin-B~\cite{liu2021swin} backbones are compared. Each backbone is evaluated under various supervision settings: LVIS~\cite{gupta2019lvis}, LVIS + IN-L~\cite{deng2009imagenet}, LVIS + IN-21k~\cite{deng2009imagenet}, and LVIS \& COCO~\cite{lin2014microsoft} + IN-21k. mAP50 is reported.}
    \resizebox{\linewidth}{!}{
    \begin{tabular}{ll|ll|ll|ll|ll|ll|ll|}
        \toprule
        &\multicolumn{1}{c}{}
        & \multicolumn{4}{c}{ResNet-50 Backbone}
        & \multicolumn{8}{c}{Swin-B Backbone} \\

        \cmidrule(lr){1-2}
        \cmidrule(lr){3-6}
        \cmidrule(lr){7-14}

        \multicolumn{2}{c}{FSOD}
        & \multicolumn{2}{|c}{LVIS} 
        & \multicolumn{2}{c|}{LVIS + IN-L}
        & \multicolumn{2}{c}{LVIS} 
        & \multicolumn{2}{c}{LVIS + IN-L} 
        & \multicolumn{2}{c}{LVIS + IN-21k} 
        & \multicolumn{2}{c}{LVIS \& COCO + IN-21k} \\

        \cmidrule(lr){3-4}
        \cmidrule(lr){5-6}
        \cmidrule(lr){7-8}
        \cmidrule(lr){9-10}
        \cmidrule(lr){11-12}
        \cmidrule(lr){13-14}

        \multirow{1}{*}{\rotatebox[origin=c]{90}{Gran}} & \multirow{1}{*}{\rotatebox[origin=c]{90}{Lev}}
         
        & \multicolumn{1}{c}{SHiNe} & \multicolumn{1}{c|}{InterHier}
        & \multicolumn{1}{c}{SHiNe} & \multicolumn{1}{c|}{InterHier}
        & \multicolumn{1}{c}{SHiNe} & \multicolumn{1}{c|}{InterHier}
        & \multicolumn{1}{c}{SHiNe} & \multicolumn{1}{c|}{InterHier}
        & \multicolumn{1}{c}{SHiNe} & \multicolumn{1}{c|}{InterHier}
        & \multicolumn{1}{c}{SHiNe} & \multicolumn{1}{c}{InterHier} \\

        \cmidrule(lr){3-4}
        \cmidrule(lr){5-6}
        \cmidrule(lr){7-8}
        \cmidrule(lr){9-10}
        \cmidrule(lr){11-12}
        \cmidrule(lr){13-14}
        
        \multirow{3}{*}{\rotatebox[origin=c]{90}{C $\xrightarrow{\hspace{0.5cm}}$ F}}
								
        & L3
        & \multicolumn{1}{c}{52.1} & \multicolumn{1}{c|}{\bf52.2({\color{higher}+0.1})}
        & \multicolumn{1}{c}{53.6} & \multicolumn{1}{c|}{\bf53.9({\color{higher}+0.3})}
        & \multicolumn{1}{c}{61.1} & \multicolumn{1}{c|}{\bf61.6({\color{higher}+0.5})}
        & \multicolumn{1}{c}{62.7} & \multicolumn{1}{c|}{\bf63.2({\color{higher}+0.5})}
        & \multicolumn{1}{c}{66.7} & \multicolumn{1}{c|}{\bf66.8({\color{higher}+0.1})}
        & \multicolumn{1}{c}{66.4} & \multicolumn{1}{c}{\bf67.1({\color{higher}+0.7})} \\
				
        & L2 
        & \multicolumn{1}{c}{39.9} & \multicolumn{1}{c|}{\bf41.0({\color{higher}+1.1})}
        & \multicolumn{1}{c}{39.8} & \multicolumn{1}{c|}{\bf41.2({\color{higher}+1.4})}
        & \multicolumn{1}{c}{46.5} & \multicolumn{1}{c|}{\bf48.2({\color{higher}+1.7})}
        & \multicolumn{1}{c}{46.6} & \multicolumn{1}{c|}{\bf48.1({\color{higher}+1.5})}
        & \multicolumn{1}{c}{51.4} & \multicolumn{1}{c|}{\bf52.1({\color{higher}+0.7})}
        & \multicolumn{1}{c}{52.4} & \multicolumn{1}{c}{\bf53.5({\color{higher}+1.1})} \\
                                            
        & L1
        & \multicolumn{1}{c}{34.3} & \multicolumn{1}{c|}{\bf36.2({\color{higher}+1.9})}
        & \multicolumn{1}{c}{31.4} & \multicolumn{1}{c|}{\bf33.5({\color{higher}+2.1})}
        & \multicolumn{1}{c}{38.5} & \multicolumn{1}{c|}{\bf42.2({\color{higher}+3.7})}
        & \multicolumn{1}{c}{35.9} & \multicolumn{1}{c|}{\bf39.2({\color{higher}+3.3})}
        & \multicolumn{1}{c}{42.2} & \multicolumn{1}{c|}{\bf44.3({\color{higher}+2.1})}
        & \multicolumn{1}{c}{42.5} & \multicolumn{1}{c}{\bf45.5({\color{higher}+3.0})} \\
         
        \bottomrule
    \end{tabular}
    }
    \label{tab:3}
\end{table*}

\subsubsection{OV-LVIS and OV-COCO}
OV-LVIS~\cite{gupta2019lvis} and OV-COCO~\cite{lin2014microsoft} are designed to evaluate the ability of a model to generalize from a set of base training classes to novel classes. OV-LVIS is a benchmark that features a large vocabulary of over 1,000 object categories with a natural, long-tail distribution of instances per category. For the open-vocabulary task, the classes are partitioned based on their frequency of occurrence. The set of 866 common and 405 frequent classes are used as the base categories for training, while the 337 rare classes are held out as the novel categories for evaluation. OV-COCO is a standard benchmark for object detection, featuring 80 common object categories found in complex, everyday scenes. For the open-vocabulary setting, the classes are partitioned into two distinct sets. These sets are composed of 48 base classes for training and 17 novel classes that are held out for testing. The evaluation is then conducted on a comprehensive validation set that includes images containing objects from both base categories and novel categories.

\subsection{Setup}

\subsubsection{Evaluation metrics}
For our primary experiments on iNatLoc~\cite{cole2022label} and FSOD~\cite{fan2020few}, we follow the cross-dataset transfer evaluation (CDTE) protocol~\cite{zhu2024survey} and use the mean average precision (mAP) at an intersection-over-union (IoU) threshold of 0.5 (mAP50). We also conduct additional experiments on OV-LVIS and OV-COCO, reporting both the overall average precision (AP) at IoU threshold of 0.5 (AP50) for these datasets. For OV-LVIS~\cite{gupta2019lvis}, we follow the open-vocabulary protocol~\cite{gu2022openvocabulary} where the rare objects are defined as novel categories. Our reported metric is the bounding box AP across IoUs from 0.5 to 0.95 for rare classes, which is denoted as AP$_r$. For OV-COCO~\cite{lin2014microsoft}, we report AP50 and evaluate InterHier performance under three distinct and comprehensive settings: (1) predict and evaluate only novel categories (Novel), (2) predict and evaluate only base categories (Base), and (3) a generalized setting to predict and evaluate all categories (All).

\subsubsection{Baseline}
We adopt SHiNe~\cite{liu2024shine} as our primary baseline method. Similar to InterHier, SHiNe is a method that focuses on improving OVD performance by integrating semantic hierarchies into text prompts. Therefore, SHiNe serves as the most relevant and direct baseline for evaluating our contributions. We use the pre-trained Detic~\cite{zhou2022detecting} as our baseline detector. Detic is a two-stage OVD detector built upon the CenterNet2~\cite{zhou2021probabilistic} framework. For classification, Detic employs an open-vocabulary classifier created using textual embeddings from the pre-trained CLIP~\cite{radford2021learning} text encoder. Classification scores for each region are then calculated through cosine similarity between the visual and the textual embeddings. Detic leverages both detection and classification datasets with image-level supervision, assigning labels to the region proposals. As backbones for this framework, we adopt a ResNet-50~\cite{he2016deep} and a Swin-B~\cite{liu2021swin} pre-trained on ImageNet-21k-P~\cite{ridnik2021imagenet}.

\subsubsection{Implementation details}
All experiments are conducted on a single NVIDIA A100 GPU environment. During the training phase, only the learnable context vectors are optimized, while the parameters of Detic~\cite{zhou2022detecting} and the pre-trained CLIP~\cite{radford2021learning} text encoder remain frozen. The number of learnable context vectors $N_l$ is set to 4 for all experiments.

\subsection{Main Results}

\begin{table}[t!]
    \centering
    \tiny
    \caption{
    Comparison of InterHier with state-of-the-art methods on OV-LVIS~\cite{gupta2019lvis}. InterHier is presented with different backbones. The table is divided into two comparison groups: ResNet-50~\cite{he2016deep} Comparison: provides a fair evaluation where all methods use the same backbone to isolate the performance of the proposed method. System-level Comparison: evaluates the overall system performance against other state-of-the-art models that utilize various backbones.}
    \resizebox{\linewidth}{!}{
    \begin{tabular}{llcc}
        \toprule
        Method & Backbone & {AP$_r$} & \textcolor{gray}{AP} \\
    
        \midrule
        \multicolumn{4}{l}{\textit{\textbf{ResNet-50 Comparison:}}} \\
        ViLD~\cite{gu2022openvocabulary} & RN50 \textcolor{gray}{(24M)} & 16.3 & \textcolor{gray}{24.4} \\
        VL-PLM~\cite{zhao2022exploiting} & RN50 \textcolor{gray}{(24M)} & 17.2 & \textcolor{gray}{27.0}\\
        OV-DETR~\cite{zang2022open} & RN50 \textcolor{gray}{(24M)} & 17.4 & \textcolor{gray}{26.6}\\
        Detic~\cite{zhou2022detecting} & RN50 \textcolor{gray}{(24M)} & 17.8 & \textcolor{gray}{26.8} \\
        F-VLM~\cite{kuo2023openvocabulary} & RN50 \textcolor{gray}{(24M)} & 18.6 & \textcolor{gray}{24.2} \\
        PromptDet~\cite{feng2022promptdet} & RN50 \textcolor{gray}{(24M)} & 19.0 & \textcolor{gray}{21.4} \\
        BARON~\cite{wu2023aligning} & RN50 \textcolor{gray}{(24M)} & 19.2 & \textcolor{gray}{26.5} \\
        MM-OVOD~\cite{kaul2023multi} & RN50 \textcolor{gray}{(24M)} & 19.3 & \textcolor{gray}{30.6} \\
        DetPro~\cite{du2022learning} & RN50 \textcolor{gray}{(24M)} & 19.8 & \textcolor{gray}{25.9} \\
        OVMR~\cite{ma2024ovmr} & RN50 \textcolor{gray}{(24M)} & 21.2 & \textcolor{gray}{30.0} \\
        OADP~\cite{wang2023object} & RN50 \textcolor{gray}{(24M)} & \underline{21.9} & \textcolor{gray}{28.7} \\
        \textbf{InterHier (Ours)} & RN50 \textcolor{gray}{(24M)} & \textbf{22.1} & \textcolor{gray}{30.2} \\
        \midrule
        
        \multicolumn{4}{l}{\textit{\textbf{System-level Comparison:}}} \\
        RegionCLIP~\cite{zhong2022regionclip} & RN50x4 \textcolor{gray}{(87M)} & 22.0 & \textcolor{gray}{32.3} \\
        CondHead~\cite{wang2023learning} & RN50x4 \textcolor{gray}{(87M)} & 24.4 & \textcolor{gray}{32.0} \\
        OWL-ViT~\cite{minderer2022simple} & ViT-L/14 \textcolor{gray}{(307M)} & 25.6 & \textcolor{gray}{34.7} \\
        F-VLM~\cite{kuo2023openvocabulary} & RN50x4 \textcolor{gray}{(87M)} & 26.3 & \textcolor{gray}{28.5} \\
        RO-ViT~\cite{kim2023region} & ViT-B/16 \textcolor{gray}{(86M)} & 28.0 & \textcolor{gray}{30.2} \\
        CORA~\cite{wu2023cora} & RN50x4 \textcolor{gray}{(87M)} & 28.1 & - \\
        CFM-ViT~\cite{kim2023contrastive} & ViT-B/16 \textcolor{gray}{(86M)} & 28.8 & \textcolor{gray}{32.0} \\
        CoDet~\cite{ma2023codet} & Swin-B \textcolor{gray}{(88M)} & 29.4 & \textcolor{gray}{39.2} \\
        DITO~\cite{kim2024region} & ViT-B/16 \textcolor{gray}{(86M)} & 32.5 & \textcolor{gray}{34.0} \\
        Detic~\cite{zhou2022detecting} & Swin-B \textcolor{gray}{(88M)} & 33.8 & \textcolor{gray}{40.7} \\
        OVMR~\cite{ma2024ovmr} & Swin-B \textcolor{gray}{(88M)} & \underline{34.4} & \textcolor{gray}{40.9} \\
        \textbf{InterHier (Ours)} & Swin-B \textcolor{gray}{(88M)} & \textbf{36.1} & \textcolor{gray}{42.6} \\
        
        \bottomrule
    \end{tabular}
    }
    \label{tab:4}
\end{table}

\subsubsection{iNatLoc}
Tab. \ref{tab:2} provides the detection performance of InterHier evaluated across six label granularity levels on iNatLoc~\cite{cole2022label}, utilizing its ground-truth hierarchical structures. To thoroughly examine the generalizability of InterHier, we compare its performance when directly integrated into baseline OVD models. This comparison is conducted with two distinct backbone architectures (e.g., ResNet-50~\cite{he2016deep} and Swin-B~\cite{liu2021swin}). Each backbone is evaluated under various supervision settings, including LVIS~\cite{gupta2019lvis}, LVIS + IN-L~\cite{deng2009imagenet}, LVIS + IN-21k~\cite{deng2009imagenet}, and LVIS \& COCO~\cite{lin2014microsoft} + IN-21k. Under all supervision and backbone settings, InterHier consistently outperforms the baseline method across all levels of label granularity. With the ResNet-50 backbone, InterHier not only demonstrates performance improvements at the coarsest granularity level (L1) but also maintains robust improvements at the finest granularity level (L6). With the Swin-B backbone, InterHier achieves superior performance across all granularity levels. In particular, InterHier excels at the coarsest granularity level (L1), with significant improvements up to 15.5\% (LVIS). These results demonstrate the effectiveness of a prepended learnable context for aligning textual and visual embeddings across diverse semantic granularities.

\subsubsection{FSOD}
Tab. \ref{tab:3} presents the performance comparison of InterHier on FSOD~\cite{fan2020few}, which is a dataset designed for few-shot object detection in an open-vocabulary setting. While InterHier yields a slight improvement on FSOD compared to the enhancement on iNatLoc in Tab. \ref{tab:2}, this difference is reflective of the less diverse hierarchical structure in FSOD. Specifically, iNatLoc provides a six-level taxonomy with rich hierarchical semantic relationships. In contrast, FSOD provides a much simpler hierarchy of three levels. This structural characteristic provides fewer hierarchical semantic relationships, which limits the advantage of InterHier that is designed to leverage rich hierarchical structures. Despite these dataset characteristics, InterHier still demonstrates consistent performance improvements on FSOD. InterHier outperforms the baseline method across various backbone architectures and supervision configurations. These results suggest that InterHier is a robust and versatile method capable of improving performance with a less diverse or limited hierarchical structure.

\begin{table}[t!]
    \centering
    \tiny
    \caption{
    Comparison of InterHier with state-of-the-art methods on OV-COCO~\cite{lin2014microsoft}. InterHier is presented with a ResNet-50~\cite{he2016deep} backbone against other state-of-the-art methods that use the same backbone architecture. Performance is reported as AP50 under three distinct settings: novel categories (Novel), base categories (Base), and all categories (All).}
    \resizebox{\linewidth}{!}{
    \begin{tabular}{llccc}
        \toprule
        \multicolumn{1}{l}{} & \multicolumn{1}{l}{} & \multicolumn{3}{c}{Generalized (17 + 48)} \\
        \multicolumn{1}{l}{\multirow{-2}{*}{Method}} & \multicolumn{1}{l}{\multirow{-2}{*}{Backbone}} & Novel & \textcolor{gray}{Base} & \textcolor{gray}{All} \\
        
        \midrule
        OVR-CNN~\cite{zareian2021open} & RN50 \textcolor{gray}{(24M)} & 22.8 & \textcolor{gray}{46.0} & \textcolor{gray}{39.9} \\
        PromptDet~\cite{feng2022promptdet} & RN50 \textcolor{gray}{(24M)} & 26.6 & - & \textcolor{gray}{50.6} \\
        RegionCLIP~\cite{zhong2022regionclip} & RN50 \textcolor{gray}{(24M)} & 26.8 & \textcolor{gray}{54.8} & \textcolor{gray}{47.5} \\
        ViLD~\cite{gu2022openvocabulary} & RN50 \textcolor{gray}{(24M)} & 27.6 & \textcolor{gray}{59.5} & \textcolor{gray}{51.2} \\
        Detic~\cite{zhou2022detecting} & RN50 \textcolor{gray}{(24M)} & 27.8 & \textcolor{gray}{51.1} & \textcolor{gray}{45.0} \\
        F-VLM~\cite{kuo2023openvocabulary} & RN50 \textcolor{gray}{(24M)} & 28.0 & - & \textcolor{gray}{39.6}\\
        OV-DETR~\cite{zang2022open} & RN50 \textcolor{gray}{(24M)} & 29.4 & \textcolor{gray}{61.0} & \textcolor{gray}{52.7}\\
        OADP~\cite{wang2023object} & RN50 \textcolor{gray}{(24M)} & 30.0 & \textcolor{gray}{53.3} & \textcolor{gray}{47.2} \\
        CoDet~\cite{ma2023codet} & RN50 \textcolor{gray}{(24M)} & 30.6 & \textcolor{gray}{52.3} & \textcolor{gray}{46.6} \\
        PB-OVD~\cite{gao2022open} & RN50 \textcolor{gray}{(24M)} & 30.8 & \textcolor{gray}{46.1} & \textcolor{gray}{42.1} \\
        VLDet~\cite{lin2023learning} & RN50 \textcolor{gray}{(24M)} & 32.0 & \textcolor{gray}{50.6} & \textcolor{gray}{45.8} \\
        MEDet~\cite{chen2022open} & RN50 \textcolor{gray}{(24M)} & 32.6 & \textcolor{gray}{54.0} & \textcolor{gray}{49.4} \\
        BARON~\cite{wu2023aligning} & RN50 \textcolor{gray}{(24M)} & 34.0 & \textcolor{gray}{60.4} & \textcolor{gray}{53.5} \\
        VL-PLM~\cite{zhao2022exploiting} & RN50 \textcolor{gray}{(24M)} & 34.4 & \textcolor{gray}{60.2} & \textcolor{gray}{53.5} \\
        CORA~\cite{wu2023cora} & RN50 \textcolor{gray}{(24M)} & 35.1 & \textcolor{gray}{35.5} & \textcolor{gray}{35.4} \\
        CCKT-Det~\cite{zhang2025cyclic} & RN50 \textcolor{gray}{(24M)} & \textbf{38.0} & - & \textcolor{gray}{35.0} \\
        \textbf{InterHier (Ours)} & RN50 \textcolor{gray}{(24M)} & \underline{37.4} & \textcolor{gray}{58.7} & \textcolor{gray}{53.1} \\
        
        \bottomrule
    \end{tabular}
    }
    \label{tab:5}
\end{table}

\begin{figure*}[t!]
    \centering
    \includegraphics[width=0.32\linewidth]{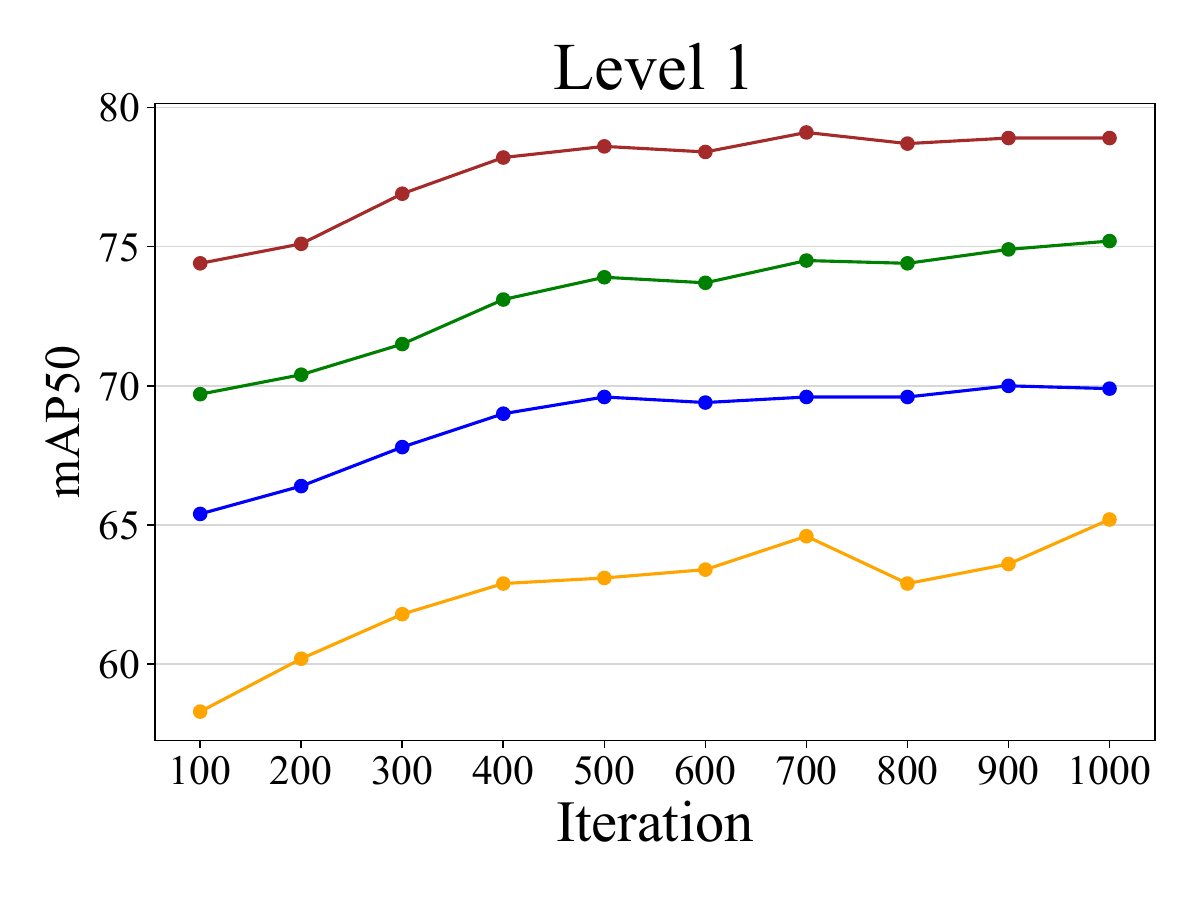}
    \hfill
    \includegraphics[width=0.32\linewidth]{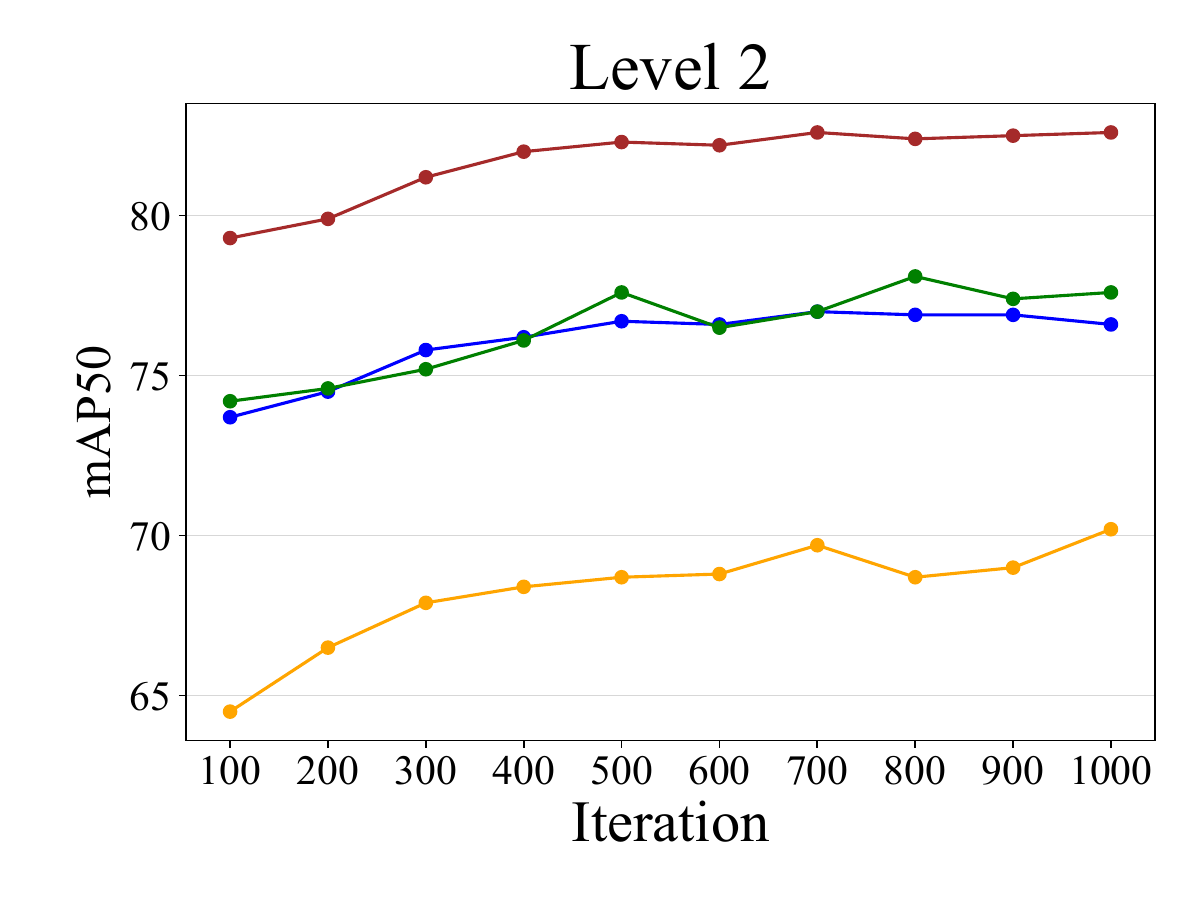}
    \hfill
    \includegraphics[width=0.32\linewidth]{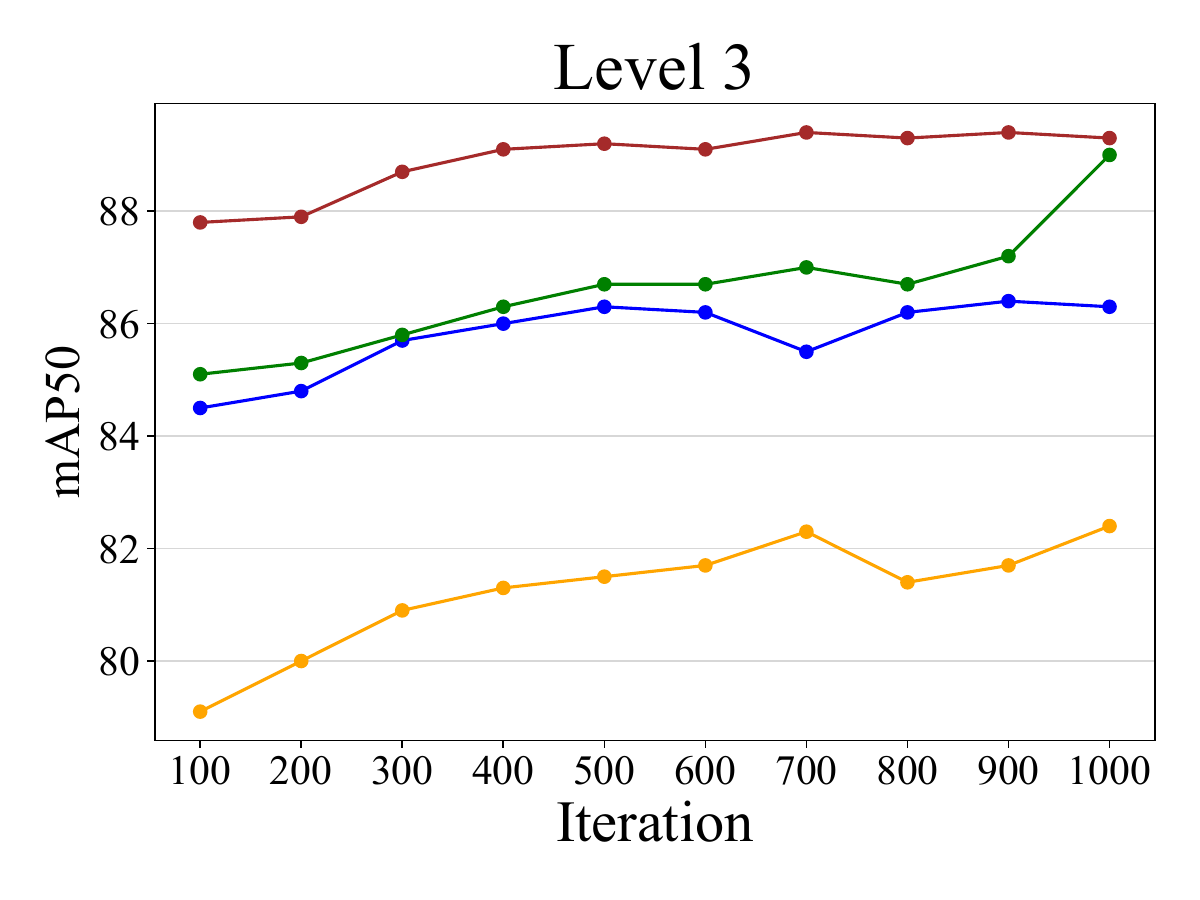}
    \\
    \includegraphics[width=0.32\linewidth]{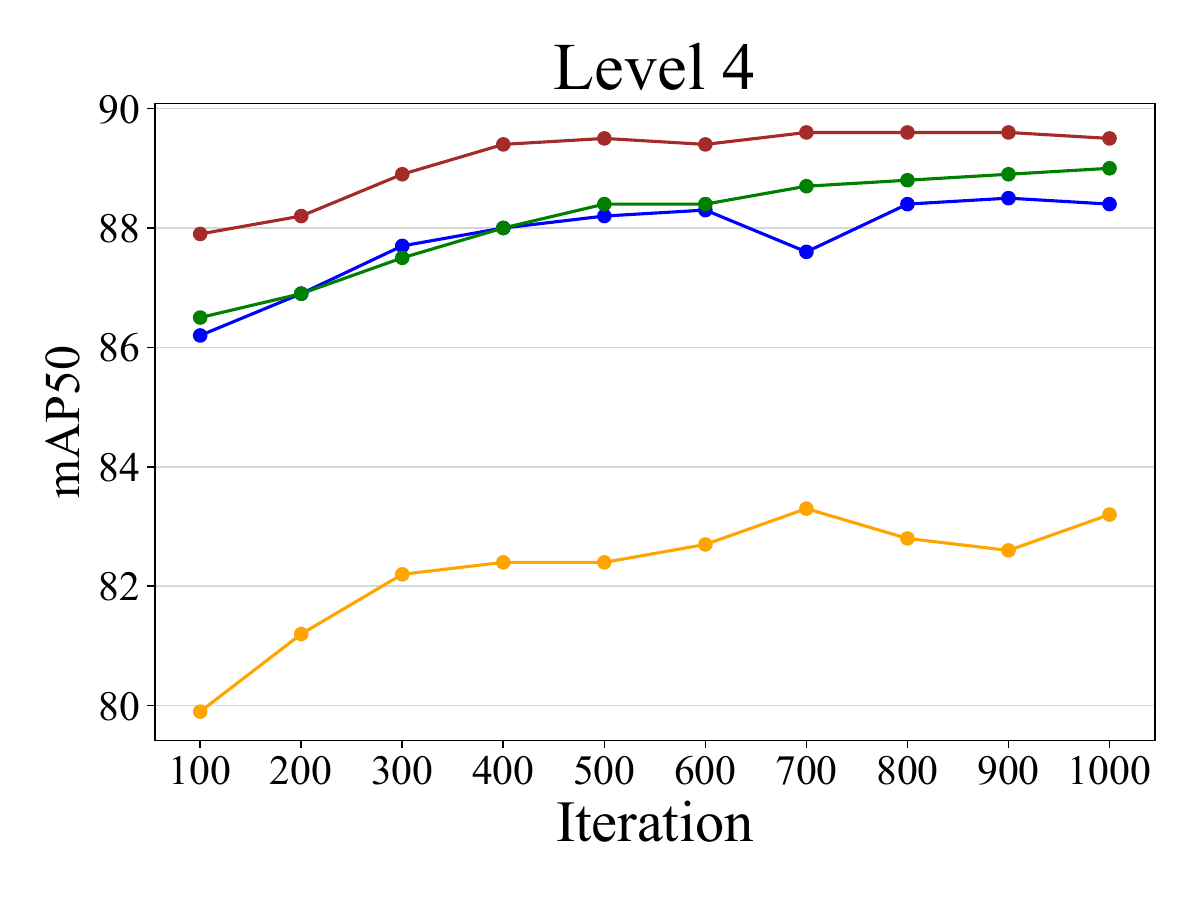}
    \hfill
    \includegraphics[width=0.32\linewidth]{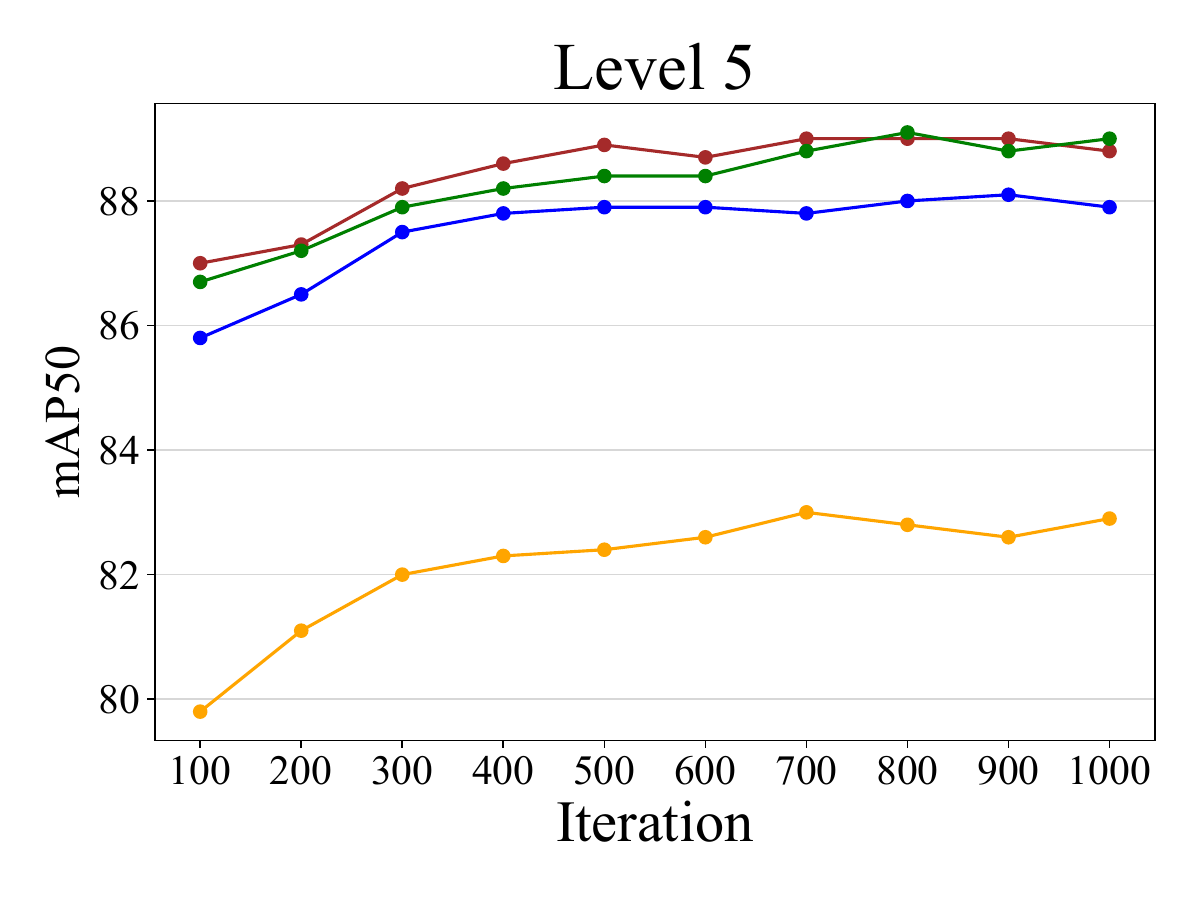}
    \hfill
    \includegraphics[width=0.32\linewidth]{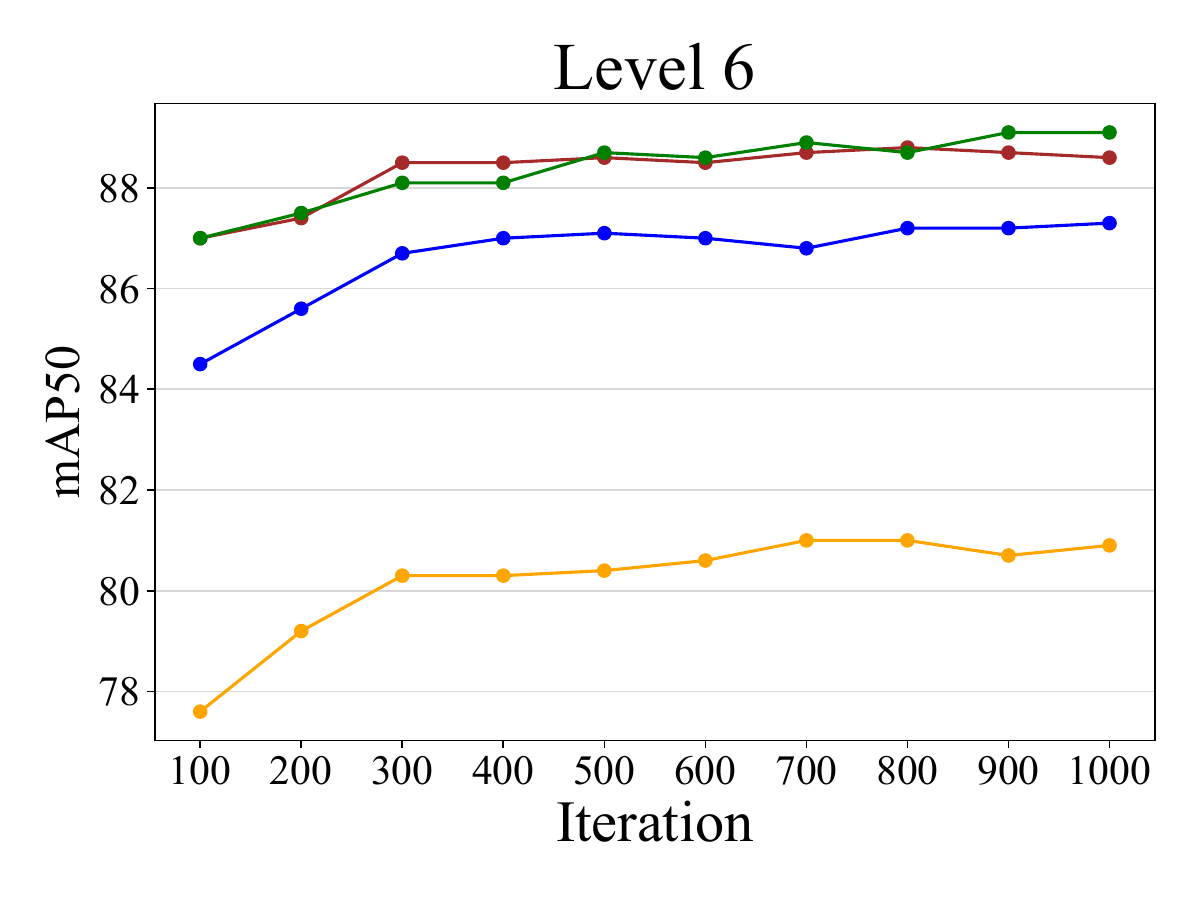}
    \\
    \begin{minipage}{\textwidth}
        \centering
        \includegraphics[width=0.6\textwidth]{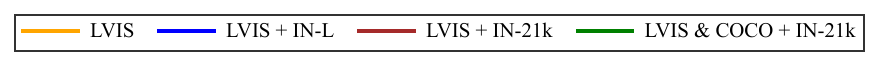}
    \end{minipage}
\caption{Performance comparison of InterHier with a Swin-B~\cite{liu2021swin} backbone evaluated at different hierarchy levels (Level 1 to Level 6). mAP50 is plotted over training iterations (100 to 1,000). Each curve represents results obtained from models trained on various datasets: LVIS~\cite{gupta2019lvis} (orange), LVIS + IN-L~\cite{deng2009imagenet} (blue), LVIS + IN-21k~\cite{deng2009imagenet} (brown), and LVIS \& COCO~\cite{lin2014microsoft} + IN-21k (green).}
\label{fig:3}
\end{figure*}

\subsubsection{OV-LVIS}
Tab. \ref{tab:4} demonstrates the comparison of InterHier with state-of-the-art methods on OV-LVIS~\cite{gupta2019lvis}. We follow a strict open-vocabulary setting~\cite{gu2022openvocabulary}, where novel categories are kept entirely unknown during the training phase. This ensures that the detector is not biased towards any specific set of novel classes. To ensure a fair comparison, Tab. \ref{tab:4} lists the backbone architecture, as this factor influences the results. Across all evaluated backbones and data settings, InterHier consistently outperforms other state-of-the-art methods. In the ResNet-50 comparison group, InterHier obtains an AP$_r$ of 22.1\%, surpassing previous methods. In the System-level comparison group, InterHier also outperforms all other state-of-the-art methods with a Swin-B backbone. InterHier achieves an AP$_r$ of 36.1\%, which is the highest score in Tab. \ref{tab:4} and exceeds other OVD models using the same backbone, such as Detic~\cite{zhou2022detecting} and OVMR~\cite{ma2024ovmr}. These results confirm that InterHier shows not only robust performance but also high generalizability for the OVD task.

\subsubsection{OV-COCO}
Tab. \ref{tab:5} presents the comparison of InterHier with state-of-the-art methods on OV-COCO~\cite{lin2014microsoft}. With a ResNet-50 backbone, InterHier achieves a competitive Novel AP50 of 37.4\%. While CCKT-Det~\cite{zhang2025cyclic} currently achieves state-of-the-art on novel classes, its performance on all categories is significantly lower. In contrast, InterHier demonstrates superior performance on base and all categories, maintaining competitive performance against other state-of-the-art methods. This performance trade-off stems from the OV-COCO dataset characteristics. If a hierarchy were constructed for OV-COCO by leveraging large language models to derive hierarchical semantics, we believe that InterHier shows improved performance on novel categories. Nonetheless, we maintained this single-level setup to ensure a fair comparison with other methods that do not employ external hierarchies on OV-COCO. Specifically, InterHier achieves a Base AP50 of 58.7\% and an All AP50 of 53.1\%. These results indicate that InterHier not only maintains competitive performance on novel classes but also demonstrates robust overall detection capabilities.

\subsection{Further Analysis}

\subsubsection{Visualization of performance convergence}

To analyze the convergence process of InterHier under different label granularities, we visualize the mAP50 performance of InterHier evaluated at diverse semantic granularity levels. Fig. ~\ref{fig:3} demonstrates the convergence trends across six hierarchical levels on iNatLoc~\cite{cole2022label}. These results are evaluated over training iterations from 100 to 1,000 with a Swin-B backbone. The evaluation is conducted with models trained on four different dataset configurations: LVIS~\cite{gupta2019lvis}, LVIS + IN-L~\cite{deng2009imagenet}, LVIS + IN-21k~\cite{deng2009imagenet}, and LVIS \& COCO~\cite{lin2014microsoft} + IN-21k. InterHier demonstrates performance improvements across all hierarchy levels throughout the training process. All models achieve rapid convergence with as few as 100 training iterations. We observe slight performance variations in several intermediate iterations. These variations are attributed to the relatively small batch size and the limited size of the training dataset. Nonetheless, such variations are temporary in nature, and the overall convergence trend continues to improve consistently throughout training. These results demonstrate that InterHier not only enhances OVD performance with minimal training but also provides rapid convergence.

\begin{figure}[!t]
    \centering
    \begin{minipage}[b]{\linewidth}
        \centering
        \includegraphics[width=0.75\linewidth]{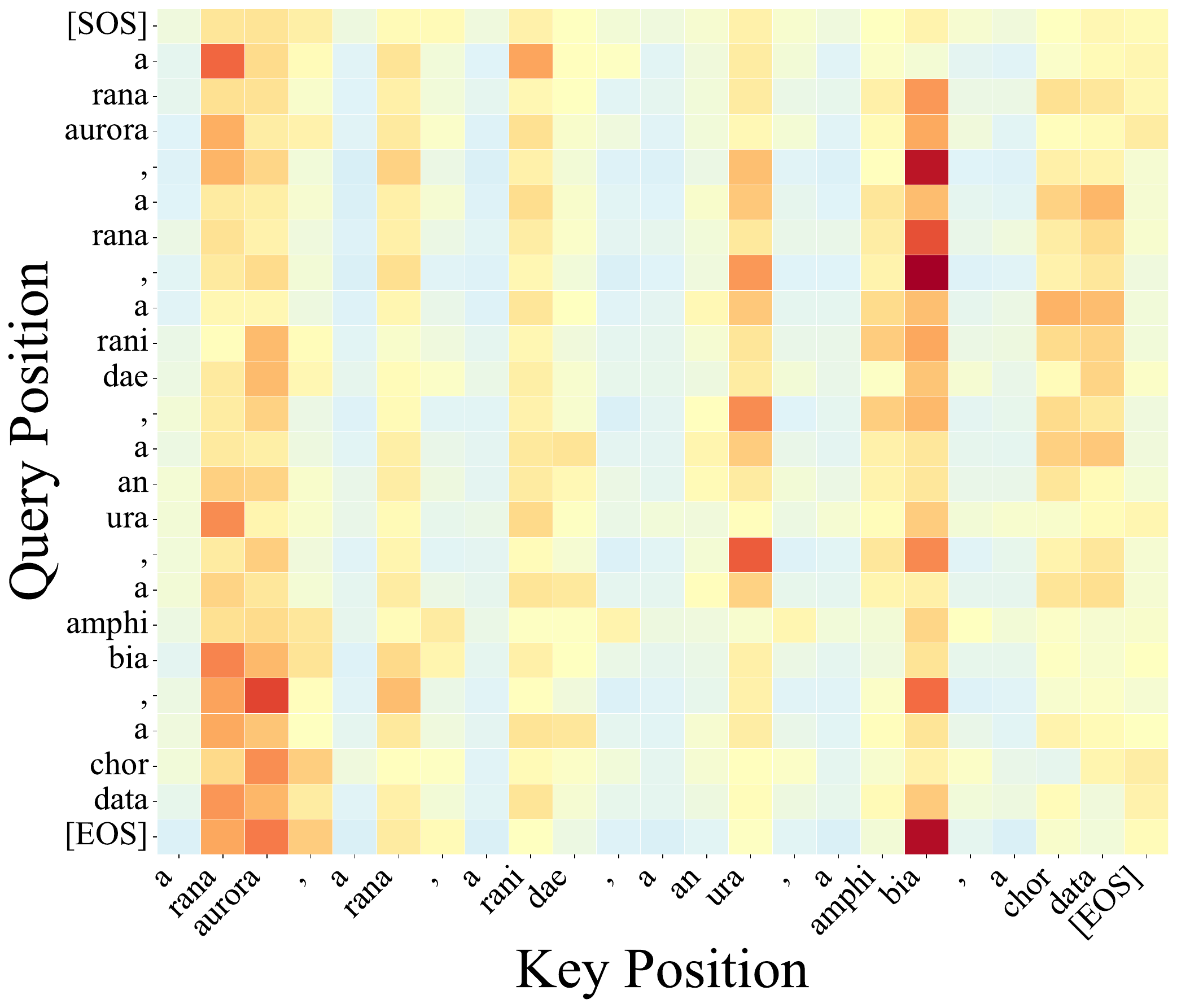}
        \centerline{(a) Baseline}
        \vspace{2.0mm}
    \end{minipage}
    \begin{minipage}[b]{\linewidth}
        \centering
        \includegraphics[width=0.75\linewidth]{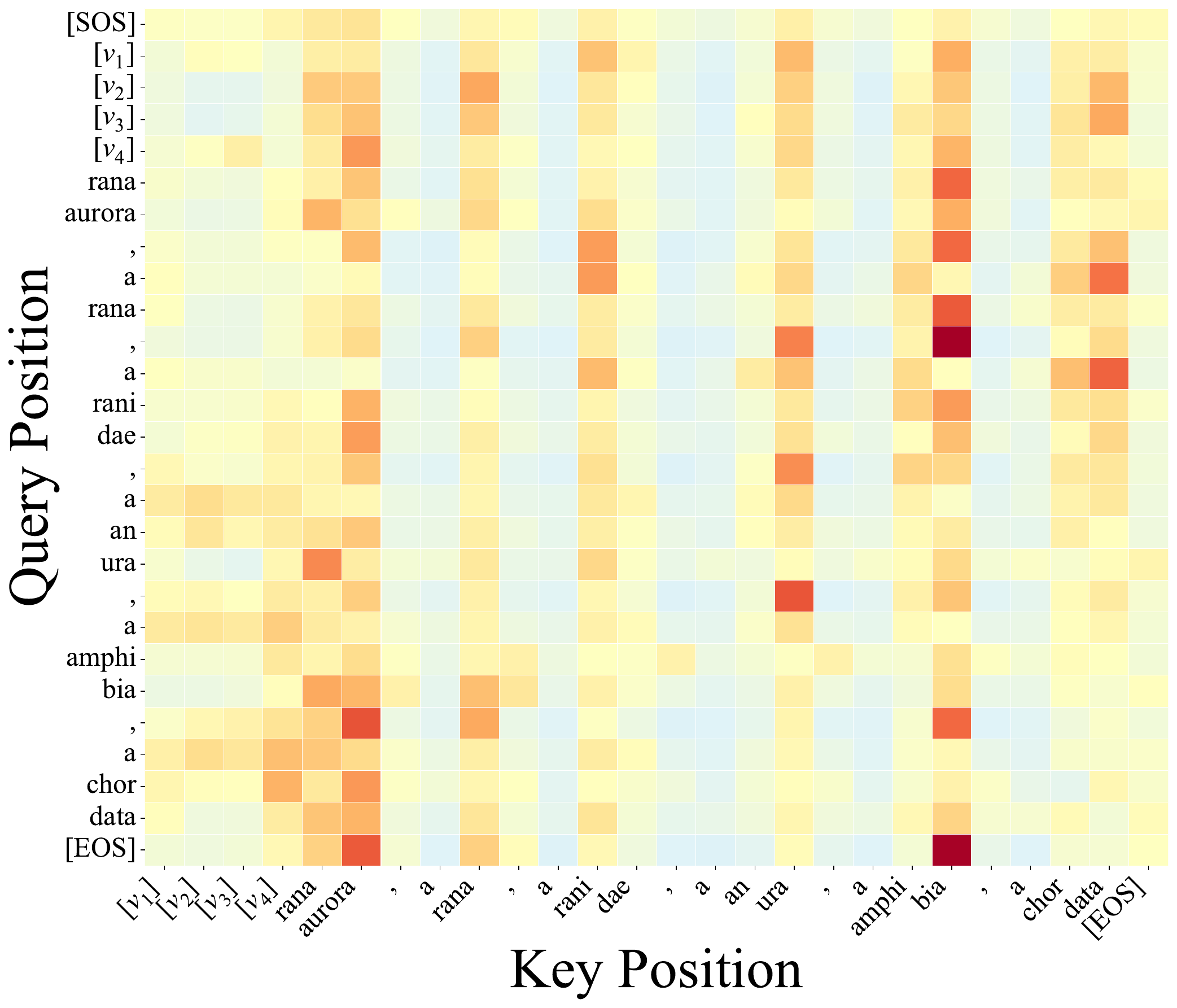}
        \centerline{(b) InterHier}
    \end{minipage}
    
    \caption{Visualization of self-attention map from the final layer of the CLIP~\cite{radford2021learning} text encoder, comparing the baseline model~\cite{zhou2022detecting} using the A connector with InterHier on iNatLoc~\cite{cole2022label}. InterHier is trained on LVIS~\cite{gupta2019lvis} \& COCO~\cite{lin2014microsoft} + IN-21k~\cite{deng2009imagenet}.}
    \label{fig:4}
\end{figure}

\subsubsection{Visualization of self-attention map}
To analyze the prepended learnable context guides the self-attention, we visualize the self-attention map from the final layer of the CLIP~\cite{radford2021learning} text encoder in Fig. ~\ref{fig:4}. We compare the self-attention patterns of the baseline model~\cite{zhou2022detecting} using the A connector against InterHier for prompts at finest granularity level (Level 6). As shown in Fig. ~\ref{fig:4}, InterHier shows a changed self-attention pattern compared to the baseline model. The prepended learnable context vectors $[v_1, v_2, v_3, v_4]$ guide the model to focus on all hierarchical levels from the finest level class ``Rana aurora'' to the coarsest level class ``Chordata''. Furthermore, InterHier redistributes the self-attention map in regions other than the learnable context. For example, the attention weight from the query position for ``Rana aurora'' to the key position for ``Chordata'' is higher under InterHier compared to the baseline model, indicating a strengthened relationship between super-/sub-categories.  These results demonstrate that the learnable context allows the model to focus on the necessary hierarchical part of the interconnected structure by adjusting the original self-attention map.

\begin{table}[t!]
    \centering
    \renewcommand{\arraystretch}{1.4}
    \caption{
    Ablation study on the learnable context connector. This table compares how mAP50 performance changes depending on how fixed connectors and learnable context are combined. The experiment is conducted using InterHier, which is built upon Detic~\cite{zhou2022detecting} with a Swin-B~\cite{liu2021swin} backbone and evaluated on iNatLoc~\cite{cole2022label}. [v] denotes the learnable context connector and the best result is highlighted in bold.}
    \resizebox{\linewidth}{!}{
    \Large
    \begin{tabular}{ccc}
        \toprule
        & Prompt & mAP50 \\
        
        \midrule
        
        \#1 & a [CLASS], which is a [CLASS], ... , which is a [CLASS] & 64.9 \\
        
        \#2 & a [CLASS], a [CLASS], ... , a [CLASS] & 68.3 \\
        
        \#3 & a [CLASS], [v] [CLASS], ... , [v] [CLASS] & 70.2 \\
        
        \#4 & [v] [CLASS], [v] [CLASS], ... , [v] [CLASS] & 69.3 \\

        \rowcolor[HTML]{EFEFEF}
        
        \#5 & [v] [CLASS], a [CLASS], ... , a [CLASS] & \textbf{75.2} \\
        
        \bottomrule
    \end{tabular}
    }
    \label{tab:6}
\end{table}

\begin{figure}[t!]
    \centering
    \includegraphics[width=0.8\linewidth]{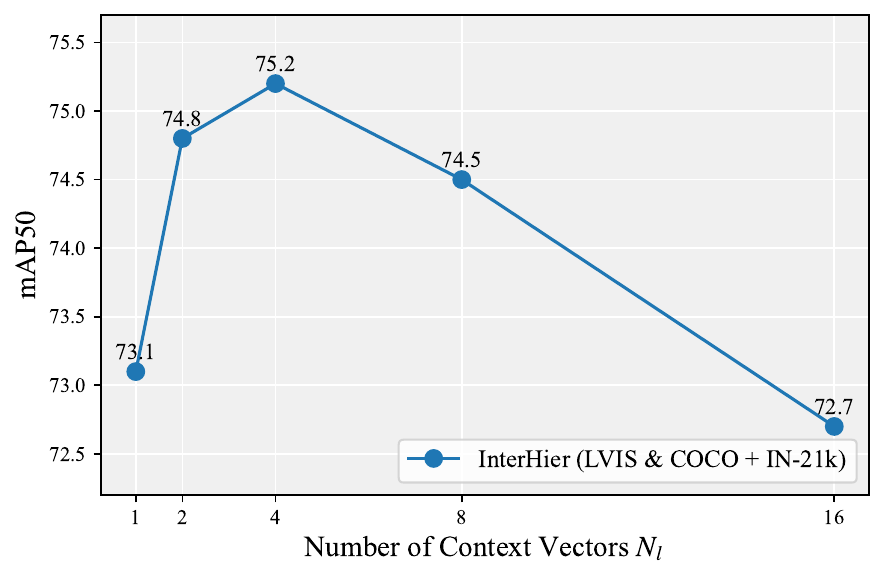}
    \caption{
    Ablation study on the number of learnable context vectors $N_l$. We report mAP50 on iNatLoc~\cite{cole2022label} as $N_l$ is ranged from 1 to 16. InterHier is trained on LVIS~\cite{gupta2019lvis} \& COCO~\cite{lin2014microsoft} + IN-21k~\cite{deng2009imagenet}.}
\label{fig:5}
\end{figure}

\begin{figure*}[!t]
    \centering
    \begin{minipage}[b]{\linewidth}
        \centering
        \includegraphics[width=\linewidth]{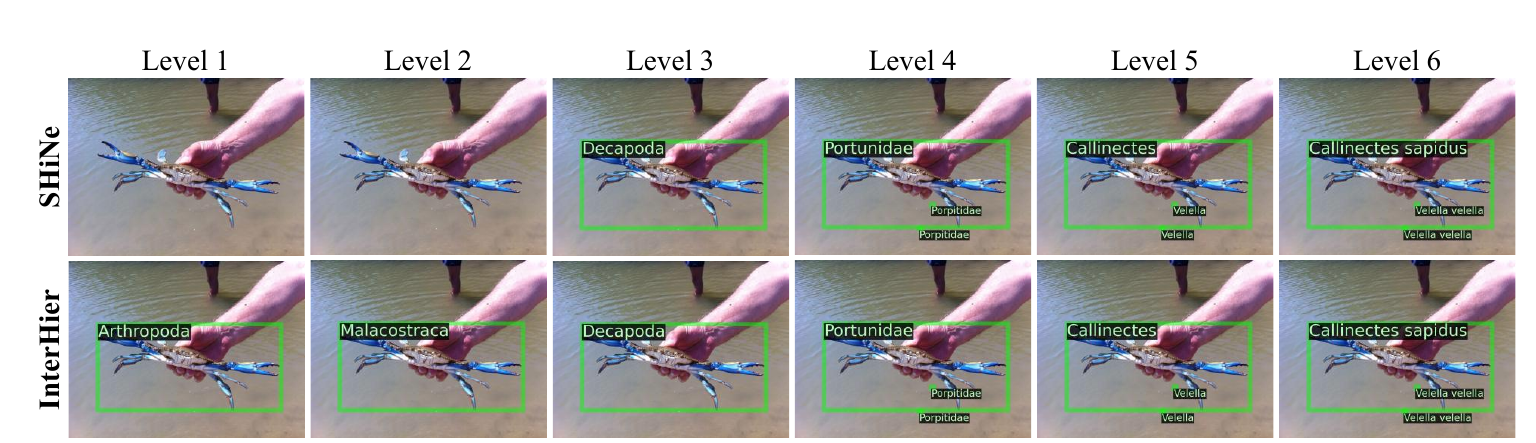}
        {(a) Qualitative results for target object ranging from ``Arthropoda'' (Level 1) to ``Callinectes sapidus'' (Level 6).}
    \end{minipage}
    
    \begin{minipage}[b]{\linewidth}
        \centering
        \includegraphics[width=\linewidth]{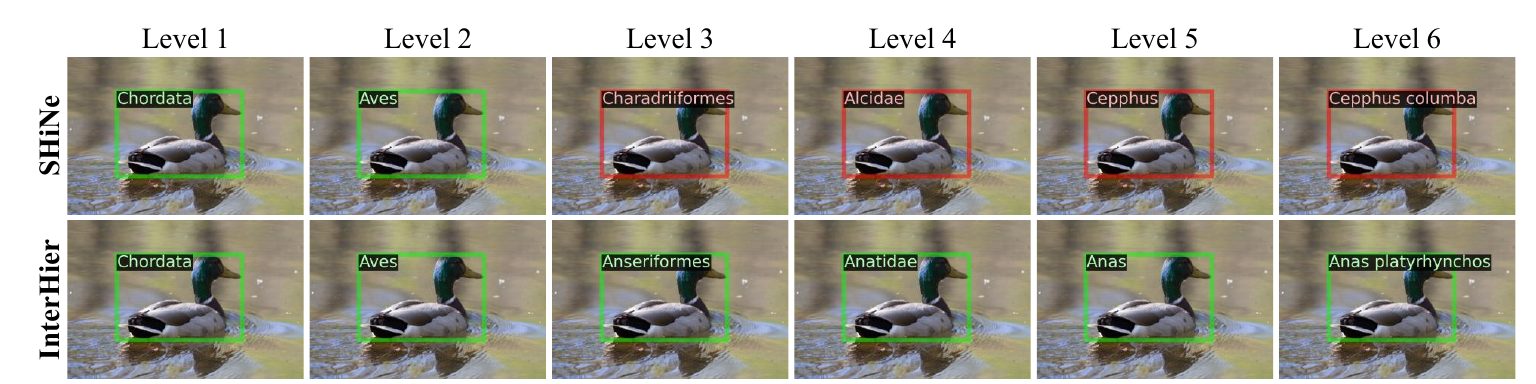}
        {(b) Qualitative results for target object ranging from ``Arthropoda'' (Level 1) to ``Callinectes sapidus'' (Level 6).}
    \end{minipage}
    
    \caption{Qualitative comparison of the detection results between SHiNe~\cite{liu2024shine} and InterHier. Both models are applied to Detic~\cite{zhou2022detecting} with a Swin-B~\cite{liu2021swin} backbone and evaluated on iNatLoc~\cite{cole2022label} across six different label granularity levels. Green boxes indicate correct classifications, while red boxes denote misclassifications. The absence of a bounding box signifies a detection failure by the model. All results are reported in terms of mAP50.}
    \label{fig:6}
\end{figure*}

\subsubsection{Ablation study on the components of the hierarchical prompt}
Tab. \ref{tab:6} presents an ablation study on the components of the hierarchical prompt, evaluating how different combinations of fixed text connectors and the learnable context affect mAP50 performance. First, we compare prompts composed of only fixed connectors (Rows \#1 and \#2). The result shows that the simpler A connector leads to a clear performance improvement over the IS-A connector. This finding suggests that for representing hierarchical relationships, simple and direct connections exhibit stronger semantic alignment than manually designed complex phrases. Subsequently, we analyze configurations that incorporate the learnable context. Replacing the fixed connectors with the learnable context (Rows \#3 and \#4) further boosts performance over the fixed connectors. The highest mAP50 of 75.2\% is achieved when the learnable context is prepended to the entire prompt, which then uses the A connector for subsequent categories (Row \#5). These results suggest that the learnable context provides superior performance when applied as a global, prepended prompt that contextualizes the entire hierarchical sequence.

\subsubsection{Ablation study on learnable context vectors}
Fig. \ref{fig:5} illustrates the results of the ablation study to determine the optimal number of learnable context vectors $N_l$ in Eq. \eqref{eq:2}. To analyze its influence, we experiment with $N_l$ from 1 to 16 and evaluate the mAP50 performance on iNatLoc~\cite{cole2022label}. InterHier is trained on LVIS~\cite{gupta2019lvis} \& COCO~\cite{lin2014microsoft} + IN-21k~\cite{deng2009imagenet}. The performance at $N_l$ of 1 is the mAP50 of 73.1\% and peaks at $N_l$ of 4, which achieves the mAP50 of 75.2\%. Increasing the number of context vectors beyond 4 leads to performance degradation, as seen with the mAP50 of 74.5\% at $N_l$ of 8 and 72.7\% at $N_l$ of 16. These results confirm that $N_l$ of 4 provides the optimal setting for our method.

\subsection{Qualitative Results}
Fig. \ref{fig:6} presents a detailed qualitative comparison of the detection results between SHiNe~\cite{liu2024shine} and our proposed InterHier. The comparison is conducted across six hierarchical levels on iNatLoc~\cite{cole2022label}. As shown in Fig. \ref{fig:4} (a), InterHier demonstrates superior detection performance over the baseline model, SHiNe, at coarser levels. For instance, at level 1, InterHier correctly classifies the object as ``Arthropoda'', while SHiNe does not provide a prediction. These results indicate that InterHier performs robustly and generalizes even at broad semantic levels, which consist of a wide and highly diverse set of objects. Furthermore, as shown in Fig. \ref{fig:4} (b), InterHier correctly detects an object at finer levels that the baseline model misclassifies. For instance, at level 6, SHiNe incorrectly classifies the bird as ``Cepphus columba'', whereas InterHier accurately identifies the object as ``Anas platyrhynchos''. These results demonstrate InterHier’s ability to distinguish subtle visual differences and to detect objects accurately at fine-grained levels, where objects are specific and visually consistent.

\section{Conclusion}
\label{sec:conclusion}

In this paper, we propose InterHier, a novel method for modeling hierarchical semantic relationships. The existing methods are not an optimal solution due to their reliance on manually designed prompts with hand-crafted connectors. In contrast, InterHier addresses the limitations of previous studies that rely on fixed connectors by prepending a global, learnable context. This context is optimized to align the resulting textual embedding of the prompt with the corresponding visual region embedding of the object. InterHier demonstrates consistently improved performance over methods that rely on fixed connectors. To this end, we validate the effectiveness of InterHier through extensive experiments on multiple OVD benchmarks. Across various backbones and supervision settings, InterHier consistently outperforms the baseline method, demonstrating its effectiveness and generalizability across diverse semantic granularities and multiple training datasets. Moreover, InterHier achieves competitive performance against other state-of-the-art methods, demonstrating its effectiveness on large-scale OVD benchmarks. InterHier offers high versatility, allowing for seamless integration into existing OVD models.

\bibliographystyle{IEEEtran}
\bibliography{REFERENCE}

\begin{IEEEbiography}[{\includegraphics[width=1in,height=1.25in,clip,keepaspectratio]{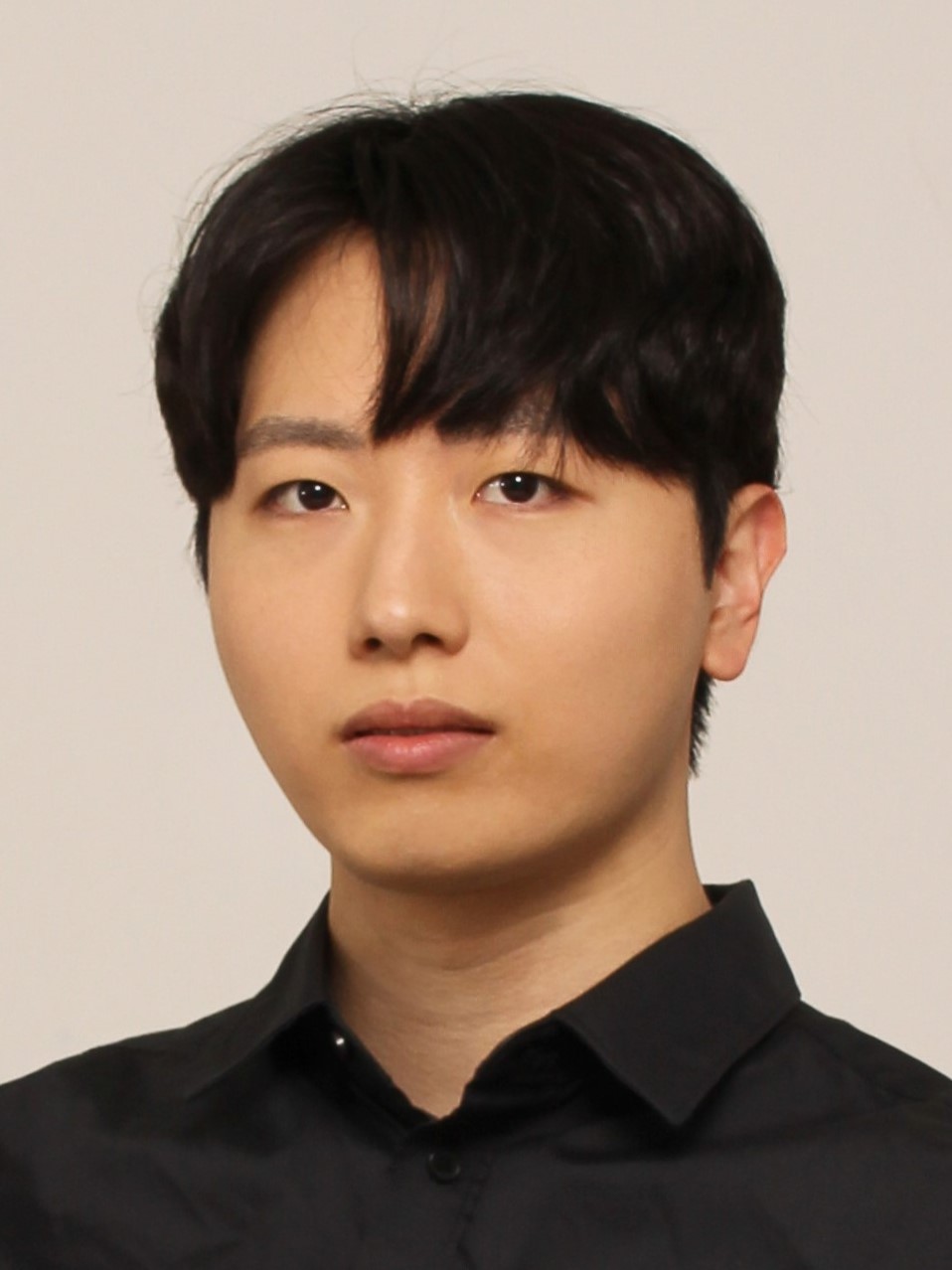}}]{Yeong-Jin Kim} received the B.S. degree in computer engineering from Hongik University, Republic of Korea, in 2024. He is currently pursuing the M.S. degree with the department of artificial intelligence, Korea University, Republic of Korea.

\end{IEEEbiography}

\begin{IEEEbiography}[{\includegraphics[width=1in,height=1.25in,clip,keepaspectratio]{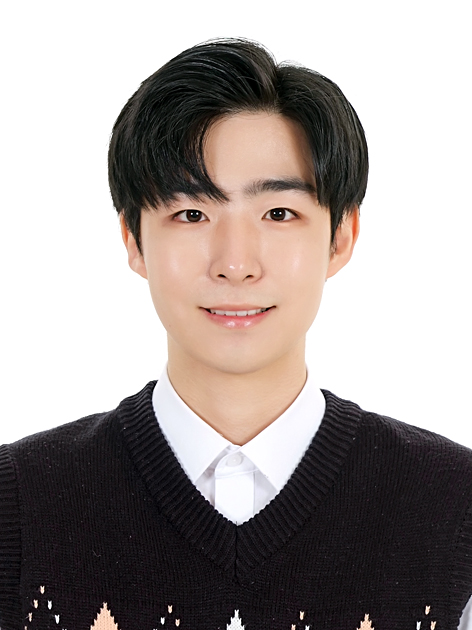}}]{Ho-Joong Kim} received the B.S. degree in the department of industrial management and engineering, and computer engineering from Hansung University, Republic of Korea, in 2021. He is currently pursuing the Ph.D. degree with the department of artificial intelligence, Korea University, Republic of Korea.
\end{IEEEbiography}

\begin{IEEEbiography}[{\includegraphics[width=1in,height=1.25in,clip,keepaspectratio]{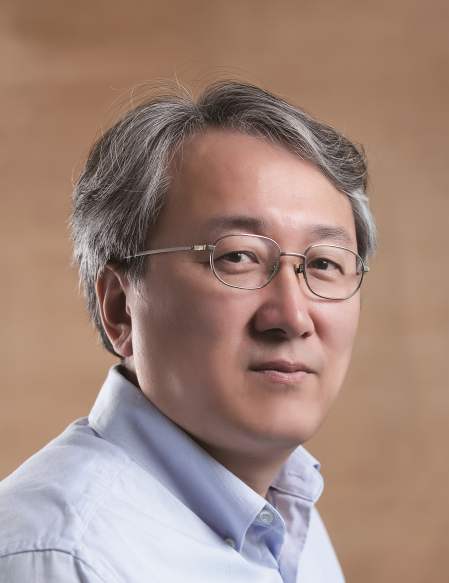}}]{Seong-Whan Lee} (Fellow, IEEE) received the B.S. degree in computer science and statistics from Seoul National University, Seoul, Republic of Korea, in 1984, and the M.S. and Ph.D. degrees in computer science from Korea Advanced Institute of Science and Technology, Seoul, Republic of Korea, in 1986 and 1989, respectively. He is currently a professor and head of the department of artificial intelligence, Korea University, Republic of Korea.
\end{IEEEbiography}

\EOD

\end{document}